\documentclass[11pt,a4paper]{article}
\usepackage[hyperref]{acl2020}
\usepackage{times}
\usepackage{latexsym}

\usepackage{amsmath,amssymb}
\usepackage{graphicx}
\usepackage{multirow}
\usepackage[utf8]{inputenc}

\usepackage{float}
\usepackage{footnote}
\usepackage{url}

\usepackage{array}
\usepackage{arydshln}
\usepackage{fdsymbol}  % card suits
\usepackage{graphicx, array, blindtext}
\usepackage{xcolor}
\usepackage{tabularx} % for \newline
\usepackage{graphbox} %loads graphicx package
\usepackage{makecell}

\newcommand{\RNum}[1]{\uppercase\expandafter{\romannumeral #1\relax}}
\usepackage[super]{nth}

\usepackage[belowskip=-6pt,aboveskip=11pt]{caption}
\usepackage[normalem]{ulem}

\definecolor{darkgreen}{rgb}{0.0, 0.5, 0.0}
\def\TBdel#1{\bgroup\markoverwith{\textcolor{darkgreen}{\rule[0.5ex]{2pt}{1pt}}}\ULon{#1}}
\def\JLdel#1{\bgroup\markoverwith{\textcolor{purple}{\rule[0.5ex]{2pt}{1pt}}}\ULon{#1}}
\def\YKdel#1{\bgroup\markoverwith{\textcolor{blue}{\rule[0.5ex]{2pt}{1pt}}}\ULon{#1}}
\def\SA#1{{\color{red}SA: \it #1}}
\def\SAdel#1{\bgroup\markoverwith{\textcolor{red}{\rule[0.5ex]{2pt}{1pt}}}\ULon{#1}}
\def\VR#1{{\color{magenta}VR: \it #1}}
\def\VRdel#1{\bgroup\markoverwith{\textcolor{magenta}{\rule[0.5ex]{2pt}{1pt}}}\ULon{#1}}

\newcommand{\ignore}[1]{}

\usepackage{enumitem}
\setlist{nolistsep}

\usepackage{microtype}

\newcommand*{\affaddr}[1]{#1} % No op here. Customize it for different styles.
\newcommand*{\affmark}[1][*]{\textsuperscript{#1}}

\aclfinalcopy % Uncomment this line for the final submission
\title{History for Visual Dialog: Do we really need it?}

\author{Shubham Agarwal\affmark[1]\thanks{\ \ This work %done during 
was carried out during the internship at Adobe Research. To appear at ACL'20.}  , Trung Bui\affmark[2], Joon-Young Lee\affmark[2], Ioannis Konstas\affmark[1] and Verena Rieser\affmark[1] \\ 
\affaddr{\affmark[1]The Interaction Lab,
Heriot-Watt University, 
Edinburgh, UK}\\
\affaddr{\affmark[2]Adobe Research, San Jose, CA, US}\\
\texttt{\{sa201, i.konstas, v.t.rieser\}@hw.ac.uk} \\
\texttt{\{bui, jolee\}@adobe.com} 
}

\date{}

\begin{document}
\maketitle
\begin{abstract}

% \VRdel{Vision and language research has seen unprecedented growth in recent years and rapidly consolidating as an interesting research arena. 
% Visual dialog has recently gained prominence which involves an agent to hold meaningful conversation with an user in the context of an image.}
%is a challenging task that 
Visual Dialog involves ``understanding'' the dialog history (what has been discussed previously) and the current question (what is asked), in addition to grounding information in the image, to %accurately 
generate the correct response. In this paper, we show that co-attention %transformer-based models
models % transformer-based  models  with  Guided-Attention  \cite{} 
%(inspired by transformers \cite{vaswani2017attention})
which explicitly encode dialog history %ical context %significantly
outperform models that don't, achieving state-of-the-art performance (72 \% NDCG on val set). However, we also expose shortcomings of the crowd-sourcing dataset collection procedure by showing that history is indeed only required for a small amount of the data and that the current evaluation metric encourages generic replies. To that end, we propose a challenging subset (VisDialConv) of the VisDial val set and provide a benchmark of 63\% NDCG.
% \SAdel{As such, we see this paper as a  plea for better datasets and task-descriptions, in order to advance the field of visual dialog.}
\end{abstract}

%%%%%%%%% BODY TEXT
\section{Introduction}

% \VRdel{
% Conversational agents which can reason about visual content is the holy grail of vision and language research. These agents have promising future to revolutionize the field and user experience in the domain of conversational image editing, video games, travel, retail and e-commerce. These systems also find potential application in VR technology, dialog based image retrieval and agents which can provide viable information to visually-impaired people about an image or other visual content. At the intersection of vision and language, Visual Dialog is an engaging task which enables users to have a conversation with a system with grounded information in visual as well as previous dialog context. This becomes a herculean task for these agents which have to first understand the context of the conversation and simultaneously ground and identify the relevant parts and objects of the image to answer it correctly. }
 
 Recently, there has been an increased interest
 in visual dialog, i.e.\ dialog-based interaction grounded in visual information
 \cite{chattopadhyay2017evaluating, de2017guesswhat,seo2017visual,guo2018dialog,shekhar2018ask,kottur2019clevr,haber2019photobook}. 
 One of the most popular test beds is the
 Visual Dialog Challenge ({\em VisDial}) \citep{das2017visual}, which involves an agent answering questions related to an image, by selecting the answer from a list of possible candidate options. According to the authors, nearly all interactions (98\%) contain dialog phenomena, such as co-reference, that can only be resolved using dialog history, which makes this a distinct task from previous Visual Question Answering (VQA) challenges, e.g. \citep{antol2015vqa}. 
 For example, in order to answer the question ``About how many?'' in Figure \ref{fig:task}, we have to infer from what was previously said, that the conversation is about the skiers.% in the picture. 
 
% \vspace{-2mm} 
% Visdial task figure
\begin{figure}[ht]
\centering
\includegraphics[scale=0.6]{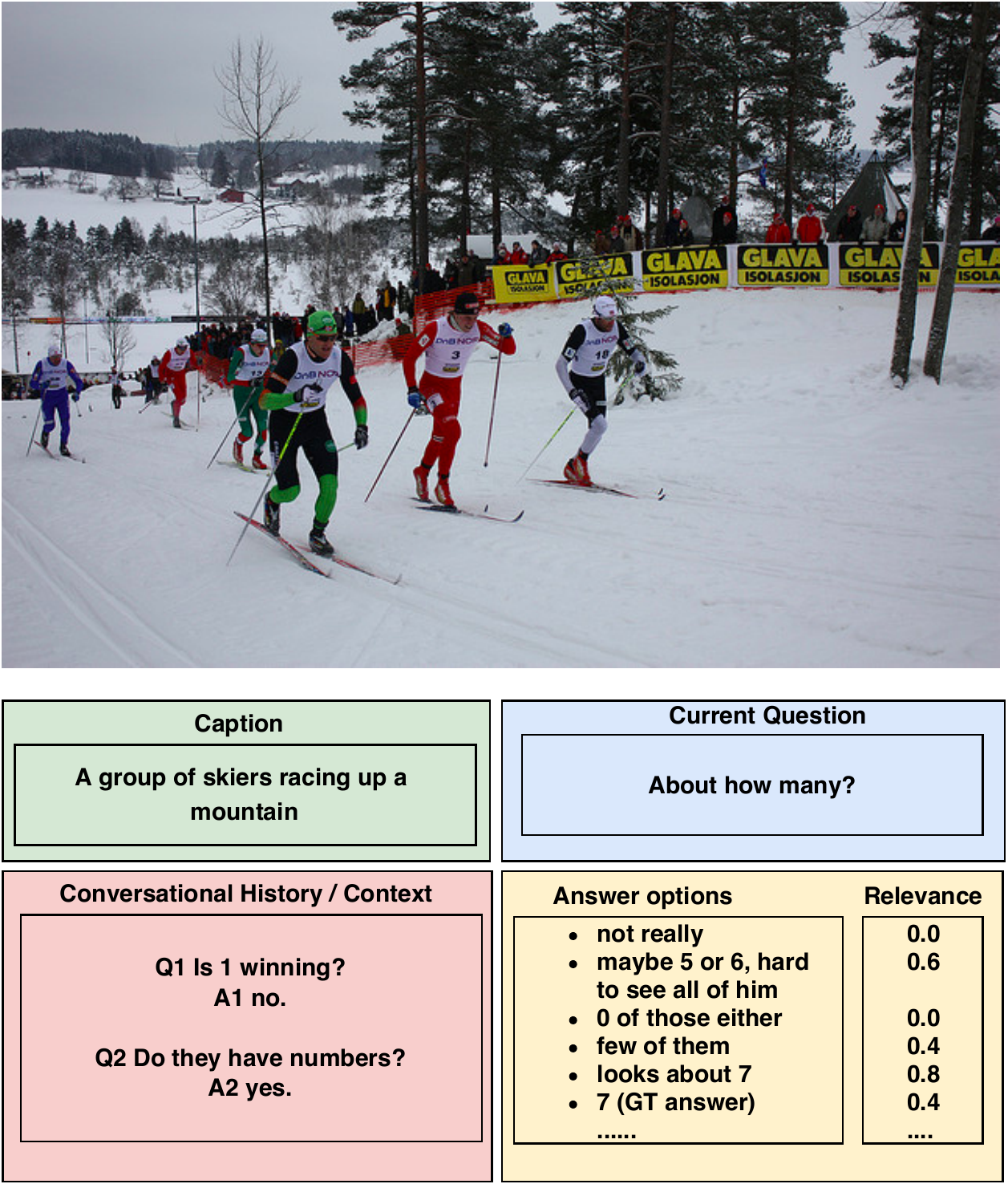}
\caption{Visual Dialog task according to \cite{das2017visual} as a ranking problem, where for the current question (blue), the agent ranks list of 100 candidate answers (yellow). Relevance weights for each candidate were collected via crowd-sourcing. Previous dialog history (red) together with the caption (green) forms the contextual information for the current turn.} 
\label{fig:task}
\end{figure}

In the original paper, \citet{das2017visual} find that models which structurally encode dialog history, such as Memory Networks \cite{bordes2016learning} or Hierarchical Recurrent Encoders \cite{serban2017hierarchical} improve performance. However, ``naive" history modelling (in this case an encoder with late fusion/concatenation of current question, image and history encodings) might actually hurt performance. \citet{massiceti2018visual} take this even further, claiming that VisDial can be modeled without taking history or even visual information into account. \citet{das2019response} rebutted by showing that both features are still needed to achieve state-of-the-art (SOTA) results and an appropriate evaluation procedure has to be used. 
 
 In this paper, we show that competitive results on VisDial can indeed be achieved by replicating the top performing model for VQA \cite{yu2019deep} 
 -- and effectively treating visual dialog as multiple rounds of question-answering, without taking history into account. %To this end, we use image encodings in combination  with deep Modular Co-Attention (MCA) Networks.
 However, we also show that these results can be significantly improved by encoding dialog history, as well as by fine-tuning on a more meaningful retrieval metric. Finally, we show that more sophisticated dialog encodings outperform naive fusion on a subset of the data which contains ``true" dialog phenomena according to crowd-workers. 
  In contrast to previous work on the VisDial dataset, e.g.\ \citep{kottur2018visual,agarwal2018ensemble,gan2019multi,guo2019image,kang2019dual}, we are the first to conduct a principled study of dialog history encodings.
 Our contributions can thus be summarized as follows:
 \begin{itemize}
     \item We present SOTA results on the VisDial dataset using transformer-based Modular Co-Attention (MCA) networks. We further show that models encoding dialog history outperform VQA models on this dataset.
     \item We show that curriculum fine-tuning ~\cite{bengio2009curriculum} on annotations of semantically equivalent answers further improves results.
    % we use weakly supervised curriculum learning ~\cite{bengio2009curriculum}
    % The basic principle behind curriculum learning is to divide a harder task into sub-problems where the learning algorithm first learns to solve simpler tasks in the form of curriculum to tackle more difficult tasks. It can be conjectured that our fine-tuning curriculum learning helps the model to generalize and select other relevant and similar answers. 
     \item We experiment with different dialog history encodings and show that early fusion, i.e.\ dense interaction with visual information (either via {\em grounding} or {\em guided attention}) %\SA{we haven't defined this yet..should we put it in `'} 
     works better for cases where conversational historical context is required.
     %information is required 
     %for grounding.
     \item We release a crowd-sourced %gold standard 
     subset containing verified dialog phenomena and provide benchmark results for future research.
 \end{itemize}

\ignore{
 Visual dialog combines recent advances in visual language grounding and multi-modal dialog research, e.g. \cite{chattopadhyay2017evaluating, de2017guesswhat,guo2018dialog, kottur2019clevr,seo2017visual}. In the following, we concentrate on the Visual Dialog (VisDial) Challenge \cite{das2017visual}, which  
 involves an agent to answer questions in the context of an image. 
 %\VR{In our work we use the Visual Dialog (VisDial) Challenge dataset \cite{das2017visual} of } 130k dialogs, each consisting of 10 QA rounds on a particular image in one session, collected using Amazon Mechanical Turk (AMT) ~\cite{buhrmester2011amazon}. 
%\VRdel{To alleviate the issue of visual priming bias and language prior, as discussed in ~\cite{zhang2016yin} for the related task of Visual Question Answering (VQA 1.0), the questioners were shown only the caption instead of the image to start their conversation. }
In contrast to preceding challenges on vision and language, such as
Visual Question Answering challenge (VQA 1.0) \cite{antol2015vqa}, VisDial has two important improvements: First, it aims to model dialog phenomena, in particular, context-dependent natural language understanding based on the dialog history. Second, it overcomes the `visual priming' bias \SA{~\cite{zhang2016yin} ?} by only showing the image to the answering agent.
%the questioners were only shown the caption instead of the image to start their conversation, in order to alleviate visual priming bias and language prior, as discussed in ~\cite{zhang2016yin} 
Thus, the questions are only grounded in the caption which serves as the initial context for the conversation. Figure \ref{fig:task} shows an example dialog for the task. For the question `does she have a bucket', the model has to understand that we are talking about the woman sitting on the bench in a park from historical context. As such, it to understand the semantics of the conversation (what has been said) and current question (what is asked) in addition to understanding the image (grounding information) to accurately generate the correct response. 
\VR{Can you change the example? I would argue that we don't necessarily need dialog history to resolve the pronoun `she', but this could also be done via common sense knowledge.}

Moving away from the limitations of BLEU and other automatic evaluation metrics for language generation tasks, as discussed for Natural Language Generation \cite{reiter2018structured, novikova2017we, agarwal2018char2char} and dialog \cite{liu2016not, ghazarian2019better, lowe2017towards}, this task has been posed as retrieval-based similar to \newcite{schatzmann2005quantitative}, where the model has to rank a list of 100 candidate answers for the current question given historical context. As the list of candidate scores is automatically generated, there can be more than one semantically correct answer, for example ``No" and ``Nope" in the figure, relevance scores for all possible candidate answers is also collected for a small subset of data, which is called ``dense annotations". 

% Visdial task figure
\begin{figure}[ht]
    \centering
    \includegraphics[scale=0.2]{pics/visdial_eg}
    \caption{Visual dialog task}    \label{fig:task}
\end{figure}

For VQA, Co-attention networks ~\cite{kim2018bilinear, nguyen2018improved} achieved significantly improved state-of-art results.
%Co-attention networks such as ~\cite{kim2018bilinear, nguyen2018improved} achieved new state-of-art on Visual Question Answering (VQA) and significantly improved the results.
%Building on top of that, 
And by extending these to deep modular co-attention networks (MCAN), \citet{yu2019deep} won the VQA Challenge 2019.  %were introduced by \newcite{yu2019deep} (challenge winners for VQA Challenge 2019). We take inspiration from their model to tackle the task of Visual Dialog which we consider as a generalization of Visual Question Answering, however, with the response also grounded on the historical context. 
Different from previous co-attention networks, they use guided attention (described in Sec. \ref{subsec:guided-attenion}) to model dense relations between the query and image regions for better visual grounding. 

In extending this to VisDial,
we use weakly supervised curriculum learning ~\cite{bengio2009curriculum}. 
The basic principle behind curriculum learning is to divide a harder task into sub-problems where the learning algorithm first learns to solve simpler tasks in the form of curriculum to tackle more difficult tasks.
As such, our model is first trained to select one correct answer based on dense annotations. On fine-tuning, it learns to identify/ rank other relevant responses using weighted cross entropy. %  We defined curriculum learning based on weighted cross entropy as a fine-tuning step. 
Cross entropy loss assigns as much probability mass as possible to the correct labels; the weighted cross entropy learns to distribute this to the other possible solutions in the candidate list. 

Our main contributions can thus be characterized as follows:
\begin{enumerate}
    \item We introduce Modular Co-Attention networks (MCA) to Visual Dialog, building on top of strong VQA models.
    %We take inspiration from strong VQA models and extend them to the task of Visual Dialog. We thus introduce a family of Modular Co-Attention networks (MCA) models to tackle this task.
    \item We formulate curriculum learning for the task of generalization to select multiple relevant options\footnote{Our curriculum learning procedure draws inspiration from the unpublished work of the Visual Dialog 2019 challenge winners}. It can be conjectured that our fine-tuning curriculum learning helps the model to generalize and select other relevant and similar answers. 
    \item We perform a detailed analysis of our results and show the limitations of our model as well as the current limitations and biases in dataset.
\end{enumerate}

The rest of the paper is organized as follows. In section \ref{sec:related work}, we discuss the related work in this field followed by our model description in section \ref{sec:method}. We then describe the dataset (in Sec. \ref{sec:dataset}) and implementation (Sec. \ref{sec:implementation}), followed by the experimental results and analysis (Sec. \ref{sec:results}) and concluding remarks in section \ref{sec:conclusion}.
} % end ignore!

%%%%%%%%%%%%%%%%%%%%%%%%%%%%%%%%%%%%%%%%%%
\section{Visual Dialog Models}
\label{sec:method}
% \VRdel{In this section we start by explaining the guided attention unit and modular co-attention layers. We will then delve later on the family of novel MCA models that we devised with their intuition behind them. }

%All our models
In this section, we extend Modular Co-Attention Networks, which won the VQA challenge 2019 \cite{yu2019deep} and adapt it to visual dialog. Different from previous co-attention networks ~\cite{kim2018bilinear, nguyen2018improved}, MCA networks use guided attention to model dense relations between the question and image regions for better visual grounding. 
In the following, we explore MCA networks with different input encodings following a `[model]-[input]' convention to refer to our MCA model variants; see Figure \ref{fig:all-models} for an overview. Whenever unspecified, images  are represented as a bag of bottom-up features, i.e.\ object level representations (see Section \ref{sec:implementation}). 

%We start this discussion first by describing the individual units in MCA module as depicted in Figure \ref{fig:mca-module}.
\begin{figure*}[ht]
\centering
\includegraphics[scale=0.5]{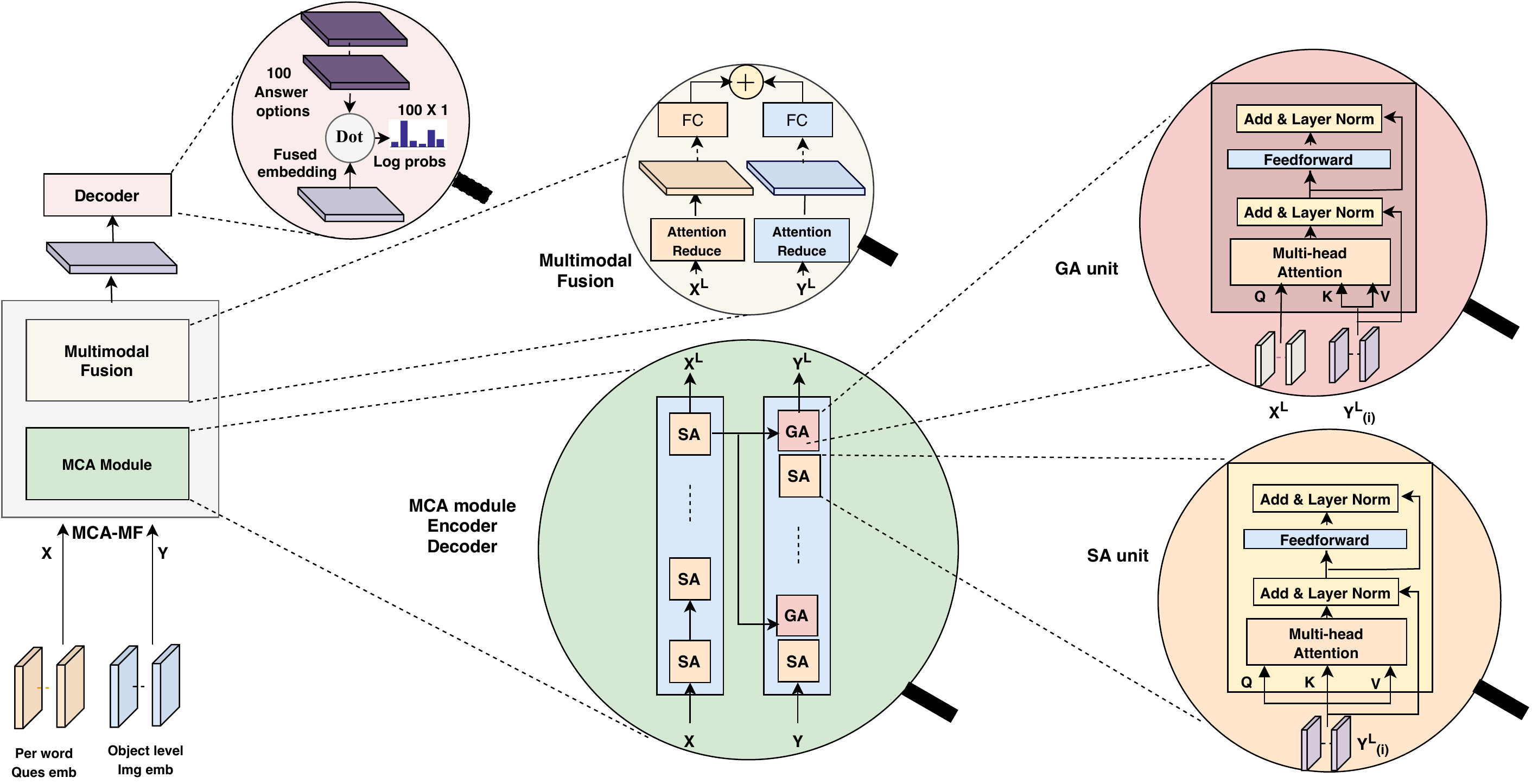}
\caption{Modular Co-Attention (MCA) module with MCA-I (Section \ref{subsec:mca-i}) %(image encoding only; see Section \ref{subsec:mca-i}) 
as an example.} % which is common to all our architectures.

\label{fig:mca-module}
\end{figure*}

%\subsection{MCA with Image component only (MCA-I)}
%\label{subsec:mca-i}

\subsection{Modular Co-Attention networks}
\label{subsec:mca-i}

%We first re-implement the original deep MCA Network introduced by \citet{yu2019deep} for the VQA task. It is based on %multiple layers of 
 
 The MCA module with multi-modal fusion % (inspired by Transformers, see \citet{yu2019deep}), 
  as depicted in Figure \ref{fig:mca-module}, is common to all our architectures. Inspired by the transformers \citep{vaswani2017attention}, the MCA network \citep{yu2019deep} is a modular composition of two basic attention units: self-attention and guided attention. These are arranged in an encoder-decoder composition in the MCA module (Figure \ref{fig:mca-module}),
% We follow the `encoder-decoder' composition in the MCA module (Figure \ref{fig:mca-module})
%  \YK{what is an encoder-decoder composition? do you mean architecture?} \SA{term \newcite{yu2019deep} used..these can be cascaded in depth as stack or encoder decoder strategy} \SA{(in the MCA module. Note our decoder is still discriminative on top of these networks)} 
 which performed best for VQA \cite{yu2019deep}. 

\subsubsection{Self-Attention and Guided-Attention}
\label{subsec:sa-ga}

The Self-Attention (SA) unit in transformers ~\cite{vaswani2017attention} is composed of a multi-head attention layer followed by a feed-forward layer. When applied to vision, the SA unit can be viewed as selecting the most relevant object-level image features for the downstream task. Specifically, the scaled dot product attention takes as input key, query and value (usually same modality's embedded representations) and outputs a self-attended vector (Eq.\ref{eqn:attn}). Multi-head attention provides multiple representation spaces to capture different linguistic/grounding phenomena, which are otherwise lost by averaging using a single head. 

\vspace{-2mm}
\begin{equation}
\small
\begin{split}
    Att(Q,K,V) &= softmax(\frac{QK^{T}}{\sqrt{d_K}})V \\ \label{eqn:attn}
    MHAtt(Q,K,V) &= Concat(head_{1},\dots head_{n})W^{O} \\
    head_{i} &= Att(QW_{i}^{Q},KW_{k}^{K},VW_{i}^{V}) 
\end{split}    
% \label{eqn:multi-head}
\end{equation}

The Guided-Attention (GA) unit conditions the attention  on different sequences. The key and value come from one  modality, while the query comes from a different modality similar to the decoder architecture in Transformers \cite{vaswani2017attention}. Similar to Eq. \ref{eqn:attn}, the GA unit outputs features  $f_{i} = Att(X, Y, Y)$
where X $\in \mathcal{R}^{m \times d_{x}}$ comes from 
one modality % the question 
and Y $\in \mathcal{R}^{n \times d_{y}}$ from the other. Residual connection ~\cite{he2016deep} and layer normalization ~\cite{ba2016layer} are applied to the output of both the attention and feed-forward layers similar to ~\citep{vaswani2017attention,yu2019deep} in both the SA and GA units.

\subsubsection{Modular Co-Attention Module}
\label{subsubsec:mca}
The following description of the MCA module is based on 
%The following applies for 
the question and the image, but can be extended analogously to model the interaction between the question and history.
% architecture for modelling dense interactions between the question and the visual representations. \cite{yu2019deep} empirically determined that initially modelling self-attention for question as well as the images separately perform better than just using guided attention on the images.
First, the %conditioning 
input (i.e.\ the question) is passed through multiple multi-head self-attention layers $L$, in order to get self-aware representations before acting as conditional signal to different modalities (visual or contextual history) similar to the auto-encoding procedure of Transformers. Then the final representation $X^{L}$ is used as the input for GA units %in the decoder 
to model cross-modal dependencies and learn the final conditioned representation $Y^{L}$. 
% \SAdel{These can be cascaded in depth to learn complex dependencies as:} \SA{The original formula IMO was for stacking..not for encoding decoding strategy.}
% \vspace{-3mm}
% \begin{equation}
% \label{eq:MCA}
%     [X^{l}, Y^{l}] = MCA^{l}([X^{l-1},Y^{l-1}])
% \end{equation}

% where $X^{0}=X$ and $Y^{0}=Y$ respectively.

\subsubsection{Multi-modal fusion}
%After the co-attention module, 
The learned representations $X^{L} \in \mathcal{R}^{m \times d}$ and $Y^{L} \in \mathcal{R}^{n \times d}$ contain the contextualized and conditioned representations over the word and image regions, respectively. We apply attention reduction \cite{yu2019deep} with a multi-layer perceptron (MLP) for $X^{L}$ (analogously for $Y^{L}$). 
We obtain the final multi-modal fused representation $z$: %is the multi-modal fused feature derived as in Eq. \ref{eq:fusion}

% For fusion fusing these representations, we follow the similar strategy as in \cite{yu2019deep} of using attention reduction to obtain a single compact representation which consists of 2 layer MLP (fc(d)-ReLU-Dropout(0.1)-fc).

\vspace{-3mm}
\begin{equation}
% \begin{align}
 \small
\begin{split}    
    \alpha^{x} = softmax(MLP^{x}(X^{L})) \\
    \widetilde{x} = \sum_{m}^{i=1} \alpha_{i}^{x}x_{i}^{L} \\
    z = LayerNorm(W_{x}^{T}\widetilde{x} + W_{y}^{T}\widetilde{y})
\end{split}
\label{eq:fusion}
% \end{align}
\end{equation}
\vspace{-2mm}

where $\alpha^{x} = [\alpha_{1}^{x} \dots \alpha_{m}^{x}] \in \mathcal{R}^{m}$ are learned attention weights (same process for $\alpha^{y}$ and $\widetilde{y}$) and $W_{x} \in \mathcal{R}^{d \times d_{z}}$, $W_{y} \in \mathcal{R}^{d \times d_{z}}$ are linear projection matrices (dimensions are the same for simplicity). 
% \TB{Explain $MLP^x$ is a Multilayer Perceptron?}

We call this model {\bf MCA with Image component only; (MCA-I)}, since it 
%This MCA network, based on \citet{yu2019deep}'s VQA model, essentially  
only encodes the question and image features %({\bf MCA with Image component only; (MCA-I)}, see Figure \ref{fig:mca-module}). It 
and therefore treats each question in Visual Dialog as an independent instance of VQA, without conditioning on the historical context of the interaction.

\begin{figure*}[ht!]
\centering
\includegraphics[scale=0.61]{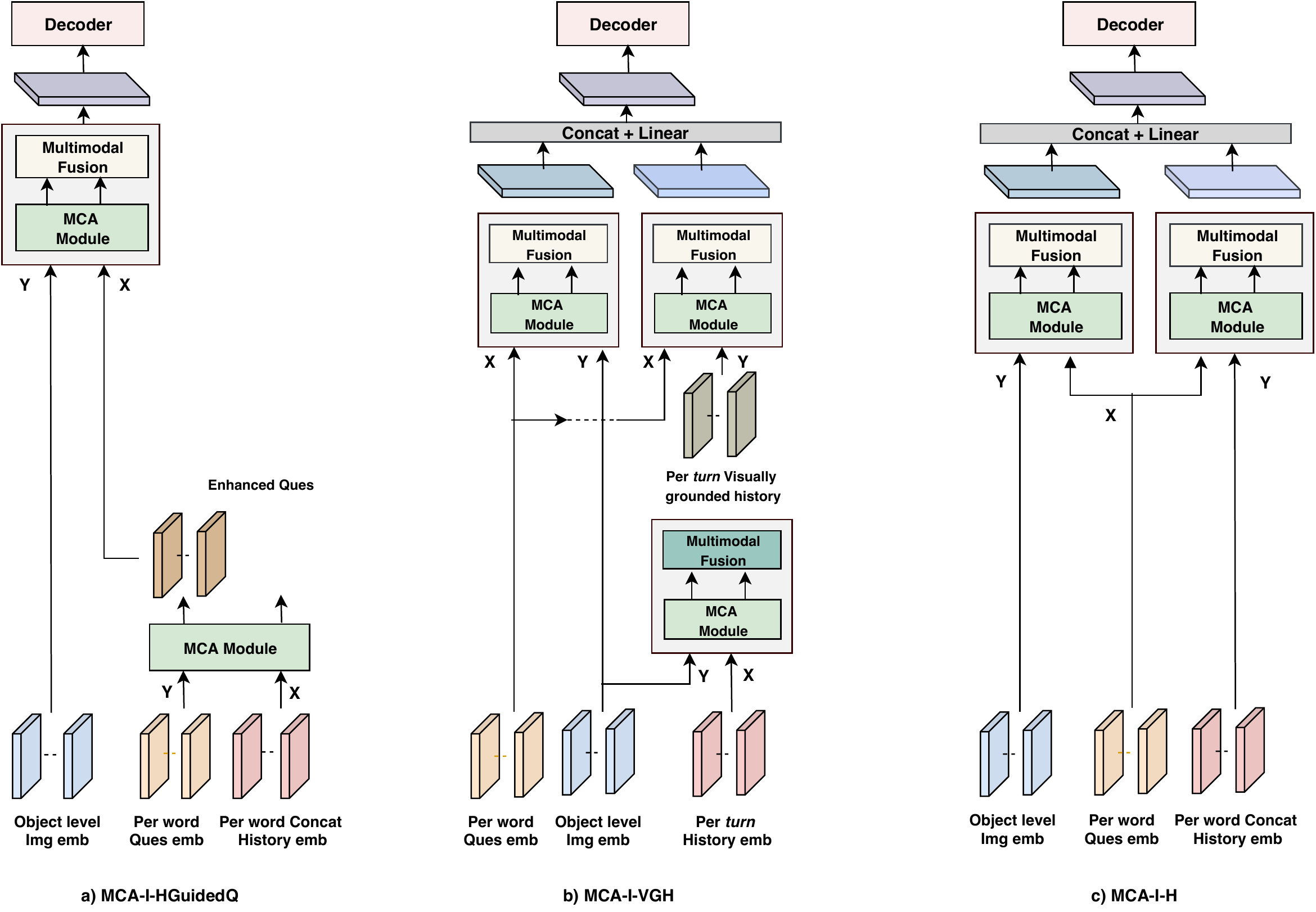}
\caption{All models incorporating dialog history described in Section \ref{subsec:hist-models}}
\label{fig:all-models}
\end{figure*}

%% SA: TODO Figure - see if this works
% https://tex.stackexchange.com/a/10247
% https://tex.stackexchange.com/a/5770

% \begin{figure*}[ht] % [!htb]
% \minipage{0.28\textwidth}%
%   \includegraphics[width=\textwidth]{pics/acl_pics/mca-i-hguidedq.pdf} \\
%   {a) MCA-I-HGuidedQ}\label{fig:mca-i-hguidedq}
% \endminipage
% \minipage{0.28\textwidth}
%   \includegraphics[width=\linewidth]{pics/acl_pics/mca-i-vgh.pdf} \\
%   {b) MCA-I-VGH}\label{fig:mca-i-vgh}
% \endminipage\hfill
% \minipage{0.28\textwidth}
%   \includegraphics[width=\linewidth]{pics/acl_pics/mca-i-h.pdf}
%   {c) MCA-I-H}\label{fig:mca-i-h}
% \endminipage\hfill
% \caption{Figure depicting all our models described in the text}
% \label{fig:all-models}
% \end{figure*}

\subsection{%MCA Network 
Variants with Dialog History}
\label{subsec:hist-models}
In the following, we extend the above framework to model dialog history. We experiment with late/shallow fusion of history and image (MCA-I-H), as well as modelling dense interaction between conversational history and the image representation (i.e.\ MCA-I-VGH, MCA-I-HGuidedQ).
% MCA-I-HConcQ, 

%\subsection{Modular co-attention network with History guided Question (MCA-I-HGuidedQ)}
\paragraph{History guided Question (MCA-I-HGuidedQ):}
\label{subsec:mca-i-hguidedq} 

The network in Figure \ref{fig:all-models}a is designed to model co-reference resolution, which can be considered as the primary task in VisDial \cite{kottur2018visual}. We first enrich the question embedding by conditioning on historical context using guided attention in the MCA module. We then use this enriched (co-reference resolved) question to model the visual interaction as described in Section \ref{subsec:mca-i}. 
%\subsection{Modular co-attention network with Visually grounded history and explicit image representation (MCA-I-VGH)}
\paragraph{Visually grounded history with image representation (MCA-I-VGH):}
\label{subsec:mca-i-vgh}

Instead of considering conversational history and the visual context as two different modalities, we now ground the history with the image first, see Figure \ref{fig:all-models}b. This is similar in spirit to maintaining a pool of visual attention maps ~\cite{seo2017visual}, where we argue that different questions in the conversation attend to different parts of the image. Specifically, we pass the history to attend to object-level image features using the MCA module to get visually grounded contextual history. We then embed the question to pool the relevant grounded history using another MCA module. In parallel, the question embedding is also used to ground the current visual context. At the final step, the respective current image and historical components are fused together and passed through a linear layer before decoding. Note, this model is generic enough %and can allow
to potentially handle multiple images in a conversation and thus could be extended for tasks e.g.\ conversational image editing, which is one of the target applications of visual dialog \cite{kim2017codraw,manuvinakurike2018edit, manuvinakurike-etal-2018-conversational,linmultimodal,elnouby2018tell_draw_repeat}.

%\subsection{Modular co-attention network with both 
\paragraph{Two-stream Image and History component (MCA-I-H):}

Figure \ref{fig:all-models}c shows the model which maintains two streams of modular co-attention networks -- one for the visual modality and the other for conversational history. We follow a similar architecture for the visual component as MCA-I and duplicate the structure for handling conversational history. At the final step, we concatenate both the embeddings and pass them through a linear layer.
%fuse by concatenation of both the embeddings.

\subsection{Decoder and loss function}

For all the models described above, we use a discriminative decoder which computes the  similarity between the fused encoding and RNN-encoded answer representations which is passed through a softmax layer to get the probability distribution over the candidate answers. We train using cross entropy over the ground truth answer: 
\vspace{-1mm}
\begin{equation}
\small
L(\theta) = \frac{1}{N} \sum_{n=1}^{N=100} y_{n}log P(x_n,\theta)
\label{eq:loss}
\end{equation}

%% SA - TODO
% \begin{figure}[ht]
% \centering
% \includegraphics[scale=0.55]{pics/acl_pics/decoder.pdf}
% \caption{Discriminative decoder}
% \label{fig:decoder}
% \end{figure}

$N$ denotes the number of candidate answers which is set to 100 for this task, $y_n$ is the (ground truth) label which is 0 or 1 during the training procedure, or a relevance score of the options during fine-tuning (casting it as multi-label classification). 
% We consider this setup weakly supervised curriculum learning since we are only using 0.16\% of training examples (with dense annotations) to fine-tune the model.

%%%%%%%%%%%%%%%%%%%%%%%%%%%%%%%%%%%%%%%%%%%%

% \vspace{-5mm}
\section{Implementation}
\label{sec:implementation}

We use PyTorch\footnote{\url{https://pytorch.org/}}~\cite{paszke2017automatic} for our experiments\footnote{Code available at \url{https://github.com/shubhamagarwal92/visdial_conv}}. Following \newcite{anderson2018bottom}, we use bottom-up features of 36 proposals from images using a Faster-RCNN ~\cite{ren2015faster} pre-trained on Visual Genome ~\cite{krishnavisualgenome} to get a bag of object-level 2048-d image representations. Input question and candidate options are tokenized to a maximum length of 20 while the conversational history  to 200. %Caption serves as context for the first question in the dialog. 
Token embeddings in text are initialized with 300-d GloVe vectors ~\cite{pennington2014glove} and shared among all text-based encoders. The RNN encodings are implemented using LSTMs ~\cite{hochreiter1997long}. We use the Adam optimizer \cite{kingma2014adam} both for training and fine-tuning. More training details can be found in Appendix \ref{sec:more-implementation}.
% SA - TODO - after submission
% \ref{sec:more-implementation}.

%%%%%%%%%%%%%%%%%%%%%%%%%%%%%%%%%%%%%%%%%%

% \vspace{-5mm}
\section{Task Description}
\label{sec:task}
% TODO - try the url 

\subsection{Dataset}\label{sec:dataset}
We use VisDial v1.0 for our experiments and evaluation.\footnote{Following the guidelines on the dataset page 
% \url{https://visualdialog.org/data}, 
we report results only on v1.0, instead of v0.9. VisDial v1.0 has been consistently used for Visual Dialog Challenge 2018 and 2019.} The dataset contains 123K/2K/8K dialogs for train/val/test set respectively. Each dialog is crowd-sourced on a different image, consisting of 10 rounds of dialog turns, totalling approx. 1.3M turns. Each question has also been paired with a list of 100 automatically generated candidate answers which the model has to rank. To account for the fact that there can be more than one semantically correct answer (e.g.\ ``Nope", ``No", ``None", ``Cannot be seen"), ``dense annotations" for 2k/2k turns of train/val of the data have been provided, i.e.\ a crowd-sourced relevance score between 0 and 1 (1 being totally relevant) for all 100 options.

\subsection{Evaluation protocol}\label{sse:eval}
As the Visual Dialog task has been posed as a ranking problem, standard information retrieval (IR) metrics are used for evaluation, such as Recall@\{1,5,10\} to measure performance in the top N results (higher better), mean reciprocal rank (MRR) of the Ground-Truth (GT) answer (higher better), and Mean rank of the GT answer (lower better). Normalized Discounted Cumulative Gain (NDCG) is another measure of ranking quality, which is commonly used when there is more than one correct answer (provided with their relevance).

\subsection{Training details}
\paragraph{Sparse Annotation Phase:}
%Since dense annotations are provided for 0.16\% of training set, 
We first train on sparse annotations, i.e.\ only 1 provided ground-truth answer, which is available for the whole training set. Here the model learns to select only one relevant answer.

\paragraph{Curriculum Fine-tuning Phase:}
Dense annotations, i.e.\ crowd-sourced relevance weights, are provided for 0.16\% of training set, which we use to fine-tune the model to select multiple semantically equivalent answers. This acts like a curriculum learning setup \cite{elman1993learning,bengio2009curriculum}, where selecting one answer using sparse annotation is an easier task and fine-tuning more difficult.\footnote{While `instance-level' curriculum learning is defined in terms of `harder dialogs', in our work, we used `dataset/task-level' curriculum finetuning. Our suggested method is a combination of curriculum learning and fine tuning (pre-training and adjusting to a specific downstream task). As such, we use the term `curriculum fine-tuning' i.e. adaptation by NDCG aware curriculum during fine-tuning.}

\subsection{Baselines}
\label{subsec:baselines}
%\VR{Please describe all the Baselines here!}
% The following baselines represent 
% \VR{pre-transformer models,
% SOTA 2019 baselines,}
% \ldots
% \YK{MCA-I, MCA-I-HConcQ and MCA-I-H too? should go here.} 

% \VR{I suggest to delete all the baselines and only keep RvA. We could even consider MCA-I as a baseline?}

\ignore{
\paragraph{\VRdel{Baseline LF with RCNN features:}}
\VRdel{We reproduced the original LF baseline model \cite{das2017visual} where the question, concatenated history and image representations are fused and subsequently passed through a linear layer and the decoder. However, the image representations consist of attended object-level representations using only the question in order to make this baseline comparable. }
}

\paragraph{MCA-I-HConcQ and MCA-H:}
\label{subsec:mca-i-hconcq}
MCA-I-HConcQ is a naive approach of concatenating raw dialog history to the question while keeping the rest of the architecture the same as MCA-I. MCA-H on the other hand considers this task as only conversational (not visual) dialog with MCA module on history instead of image.

\ignore{
\paragraph{\VRdel{OnlyQA:}}
\VRdel{In the spirit of \citet{massiceti2018visual} we develop a simple baseline %(although not statistical \YK{what does that mean? Is it Rule-based?}) 
where we recast the whole task as QA using only question embeddings; denoted as OnlyQA (w/o I w/o H) in Table \ref{table:val-curriculum}.
}
}

\paragraph{RvA:}
We reproduce the results of \citet{niu2019recursive}'s Recursive Visual Attention model (RvA), which won the 2019 VisDial challenge. %in order to demonstrate state-of-the-art results. 
Their model browses the dialog history and updates the visual attention recursively until the model has sufficient confidence to perform visual co-reference resolution.
We use their single model's open-source implementation and apply our fine-tuning procedure on the val set in Table \ref{table:val-curriculum}.
%\footnote{We consider RvAastate-of-art since they were the challenge winners of Visual Dialog Challenge 2019. It can be argued that probably the ensembling got them the bigger uplift compared to other models; which we don't deny but since our curriculum learning is inspired by their two stage procedure (for which the exact details are not published); we go with their model to consider them SOTA to have a fair comparison with their architecture.}
When reporting on the test set results in Table \ref{table:results-test}, we use the leaderboard scores published online which contains further unpublished enhancements based on ensembling (MReaL-BDAI).
% \VR{What's MS-D365-AI???}

%%%%%%%%%%%%%%%%%%%%%%%%%%%%%%%%%%%%
%\section{Experiments, Results and Analysis}
\section{Results}
\label{sec:results}

%%% Note: First tables then text in latex for easier writing

%% Table dev set
\begin{table*}[ht]
\centering
\resizebox{\textwidth}{!}{
\begin{tabular}{lcccccc|cccccc}
\hline
\multicolumn{1}{l}{\multirow{2}{*}{\textbf{Model}}} & \multicolumn{6}{c}{\textbf{Sparse annotation Phase}} & \multicolumn{6}{c}{\textbf{Curriculum Fine-tuning}} \\ \cline{2-13} 
\multicolumn{1}{l}{}                       & \multicolumn{1}{c}{\textbf{NDCG $\uparrow$}} & \multicolumn{1}{c}{\textbf{MRR $\uparrow$}} & \multicolumn{1}{c}{\textbf{R@1 $\uparrow$}} & \multicolumn{1}{c}{\textbf{R@5 $\uparrow$}} & \multicolumn{1}{c}{\textbf{R@10 $\uparrow$}} & \multicolumn{1}{c}{\textbf{Mean $\downarrow$}} & \multicolumn{1}{c}{\textbf{NDCG $\uparrow$}} & \multicolumn{1}{c}{\textbf{MRR $\uparrow$}} & \multicolumn{1}{c}{\textbf{R@1 $\uparrow$}} & \multicolumn{1}{c}{\textbf{R@5 $\uparrow$}} & \multicolumn{1}{c}{\textbf{R@10 $\uparrow$}} & \multicolumn{1}{c}{\textbf{Mean $\downarrow$}} \\ 
\hline \hline
% Baseline LF & - & 58.07 & 43.82 & 74.68 & 84.07 & 5.78 \\
% Baseline LF (Reproduced with Faster RCNN features) & 52.23 & 60.47 & 46.36 & 77.58 & 87.06 & 4.90 \\
% Only QA (w/o I w/o H) & 47.46 & 51.58 & 37.11 & 68.07 & 78.68 & 7.30 \\
% VGH-LF & 54.47 & 61.42 & 47.32 & 78.81 & 88.18 & 4.62 \\
RvA (Challenge winners; single model) & 55.86 & 64.42 & 50.71 & 81.50 & 90.15 & 4.06 &
67.90 & 51.92 & 36.57 & 70.69 & 83.61 & 5.85 \\
\hdashline[0.5pt/2pt]
MCA-H & 51.67 & 59.65 & 45.21 & 77.01 & 86.79 & 4.92 & 64.06 & 38.16 & 22.86 & 54.99 & 71.24 & 9.19 \\
MCA-I & 59.94 & 59.67 & 45.95 & 76.15 & 86.24 & 5.24 & 70.82 & 37.34 & 21.22 & 56.13 & 72.74 & 9.23 \\
MCA-I-HConcQ & 60.65 & 64.08 & 50.83 & 80.74 & 89.62 & 4.22 & 70.81 & 40.75 & 24.53 & 60 & 75.11 & 8.13 \\
\hdashline[0.5pt/2pt]
MCA-I-HGuidedQ & 60.17 & 64.36 & 50.99 & 80.95 & 89.93 & 4.17 & 71.32 & 44.1 & 28.44 & 61.74 & 76.53 & 7.83 \\
MCA-I-VGH & \textbf{62.44}%$^\spadesuit$
& 61.25 & 47.5 & 78.16 & 87.8 & 4.74 & 72.0 & 40.22 & 24.38 & 58.8 & 73.77 & 8.44 \\
\textbf{MCA-I-H} & 60.27 & \textbf{64.33} & \textbf{51.12} & 80.91 & 89.65 & 4.24 & \textbf{72.22} & 42.38 & 26.94 & 60.17 & 75.2 & 8.2 \\
\hdashline[0.5pt/2pt]
MCA-I-H-GT & 60.27 & 64.33 & 51.12 & 80.91 & 89.65 & 4.24 & 72.18 & 46.92 & 32.09 & 63.85 & 78.06 & 7.37\\
\hline
\end{tabular}
}
\caption{Results on VisDial v1.0 val set. Here `I' denotes image modality while `H' refers to the use of dialog history. Our baseline models are defined in Section \ref{subsec:mca-i} and \ref{subsec:baselines}. MCA variants with dialog history follow the same order as Section \ref{subsec:hist-models}. MCA-I-H-GT refers to the model with corrected dense annotations (see Section \ref{subsec:ndcg})
}
\label{table:val-curriculum}
\end{table*}

% Table test set
\begin{table}[ht]
\centering
\resizebox{0.48\textwidth}{!}{
\begin{tabular}{lcccccc}
\hline
\textbf{Model} & \textbf{NDCG $\uparrow$} & \textbf{MRR $\uparrow$} & \textbf{R@1 $\uparrow$} & \textbf{R@5 $\uparrow$} & \textbf{R@10 $\uparrow$} & \textbf{Mean $\downarrow$}  \\
\hline \hline
RvA & 55.59 & 63.03 & 49.03 & 80.40 & 89.83 & 4.18 \\
MS-D365-AI (Ensemble-\nth{2}) & 64.78 & 54.23 & 42.88 & 65.38 & 76.12 & 6.50 \\
% \RNum{1}
MReaL-BDAI (Ensemble-\nth{1}) & 74.57 & 53.37 & 40.96 & 66.45 & 79.70 & 6.60 \\
\hdashline[0.5pt/2pt]
MCA-I & 70.97 & 35.65 & 19.32 & 54.57 & 71.39 & 9.51 \\
MCA-I-VGH & 71.33 & 38.92 & 22.35 & 58.42 & 74.5 & 8.69 \\
\textbf{MCA-I-H} & \textbf{72.47} & 37.68 & 20.67 & 56.67 & 72.12 & 8.89 \\
\hline
\end{tabular}
}
\caption{%Automatic evaluation based on \textsc{NDCG}, \textsc{Recall@\{1,5,10\}}, \textsc{MRR} and Mean rank 
Evaluation on test-std set with results from the online leaderboard. Winners are picked on NDCG. MReaL-BDAI (2019 winning entry) is an ensemble of 6 RvA models. %Our \textbf{single model} MCA-I-H is over the rest of other submissions. 
Runner-up MS-D365AI (unpublished) also used ensembling. Note all our submitted MCA models use curriculum fine-tuning and no ensembling.}
%I/H denotes usage of visual/conversational modality.}
\label{table:results-test}
\end{table}

% Table: subsets of dataset
\begin{table*}[ht]
\centering
\resizebox{\textwidth}{!}{
\begin{tabular}{lcccccc|cccccc}
% \hline
% \multicolumn{1}{l}{\textbf{VisPro subset dataset}}  \\
\hline
\multicolumn{1}{l}{\multirow{2}{*}{\textbf{Model}}} & \multicolumn{6}{c}{\textbf{Sparse annotation Phase}} & \multicolumn{6}{c}{\textbf{Curriculum Fine-tuning}} \\ \cline{2-13} 
\multicolumn{1}{l}{}                       & \multicolumn{1}{c}{\textbf{NDCG $\uparrow$}} & \multicolumn{1}{c}{\textbf{MRR $\uparrow$}} & \multicolumn{1}{c}{\textbf{R@1 $\uparrow$}} & \multicolumn{1}{c}{\textbf{R@5 $\uparrow$}} & \multicolumn{1}{c}{\textbf{R@10 $\uparrow$}} & \multicolumn{1}{c}{\textbf{Mean $\downarrow$}} & 
\multicolumn{1}{c}{\textbf{NDCG $\uparrow$}} & \multicolumn{1}{c}{\textbf{MRR $\uparrow$}} & \multicolumn{1}{c}{\textbf{R@1 $\uparrow$}} & \multicolumn{1}{c}{\textbf{R@5 $\uparrow$}} & \multicolumn{1}{c}{\textbf{R@10 $\uparrow$}} & \multicolumn{1}{c}{\textbf{Mean $\downarrow$}} \\
\hline \hline
\textbf{VisPro subset dataset} \\
\hline
MCA-I & 59.80 & 57.88 & 45.39 & 72.24 & 82.76 & 5.84 & 69.82 & 36.2 & 20 & 54.08 & 70.92 & 10.02 \\
MCA-I-HConcQ & 61.08 & \textbf{61.79} & \textbf{48.95} & \textbf{77.5} & \textbf{86.58} & \textbf{4.72} & 68.44 & 38 & 22.24 & 55.79 & 71.71 & 9.17 \\
MCA-I-HGuidedQ & 61.35 & 60.13 & 47.11 & 75.26 & 86.18 & 5.23 & 68.29 & 36.59 & 21.05 & 53.29 & 70.13 & 9.76 \\
MCA-I-VGH & 61.68 & 59.33 & 46.18 & 75.53 & 86.71 & 5.07 & 68.97 & 39.21 & 23.68 & \textbf{57.11} & \textbf{70.53} & \textbf{8.83} \\
\textbf{MCA-I-H} & \textbf{61.72} & 59.62 & 45.92 & 77.11 & 86.45 & 4.85 & \textbf{70.87} & \textbf{39.8} & \textbf{25.39} & 55.13 & 70.39 & 9.42 \\
\hline
\textbf{VisDialConv (Crowd-sourced subset) dataset} \\
\hline
MCA-I & 52.07 & 55.55 & 41.65 & 72.47 & 83.81 & 5.92 & 58.65 & 36.2 & 20.52 & 53.3 & 68.25 & 10.32 \\
MCA-I-HConcQ & 54.84 & 62.06 & 47.42 & \textbf{80.1} & \textbf{88.87} & \textbf{4.37} & 61.42 & 37.92 & 21.86 & 55.67 & \textbf{73.3} & \textbf{9.01} \\
MCA-I-HGuidedQ & 53.81 & \textbf{62.29} & \textbf{48.35} & \textbf{80.1} & 88.76 & 4.42 & \textbf{62.92} & 38.07 & 22.58 & 54.74 & 70.82 & 9.5 \\
MCA-I-VGH & \textbf{55.48} & 58.45 & 44.54 & 74.95 & 86.19 & 5.18 & 60.63 & 38.1 & 22.89 & 53.71 & 70.31 & 9.49 \\
\textbf{MCA-I-H} & 53.01 & 61.24 & 47.63 & 79.07 & 87.94 & 4.77 & 59.89 & \textbf{39.73} & \textbf{25.15} & \textbf{56.49} & 71.86 & 9.53 \\
\hline
\end{tabular}
}
\caption{Automatic evaluation on the subsets of VisPro and VisDialConv dataset. We found history based MCA models to outperform significantly compared to the MCA-I model. On VisDialConv, MCA-I-VGH still outperform all other models in spare annotation phase while MCA-I-HGuidedQ performs the best after fine-tuning.}
\label{table:val-subset-data}
\end{table*}

%%% Text starts here 
In the following, we report results on the VisDial v1.0 val set, (Table \ref{table:val-curriculum}),
as well as the test-std set\footnote{We only report results for our best preforming models as the number of allowed submissions to the challenge is limited.},
%provided by the challenge organisers 
 (Table \ref{table:results-test}). For measuring significance (reported on $p\leq0.05$), we use Kruskal-Wallis \cite{kruskal1952use} and Wilcoxon signed rank test \cite{wilcoxon1992individual} with Bonferroni correction \cite{bonferroni1936teoria}.
We report results in terms of NDCG, which is the main metric of the challenge.
%Our results show:
%\begin{itemize}[noitemsep]
   % \item Transformer-based \SA{(co-attention)?} models outperform SOTA 2019 models. 
  % \item  %In partcular, MCA-I-VGH outperforms RvA by 6 NDCG points on the sparse annotation data and by MCA-I-H improves by 4 points over RvA on the dense annotation dev set. 
   
  %  \item Compared to MCA-I, encoding history improves results, but only significantly for MCA-I-VGH 
 %   in the sparse annotation phase. After fine-tuning MCA-I-VGH and MCA-I-H perform equally.
    %Main improvement comes from MCA/ transformer-like co-attention. 
   % \item Best history encoding? Are history encoding performing significantly different?
   % \item Curriculum fine-tuning generally significantly improves performance with an average improvement of 11.7  NDCG points, but worsens performance in terms of other retrieval-based metrics.
%\end{itemize}

MCA-I-H is our best performing model. It achieves state-of-the-art performance: It outperforms the RvA baseline by almost 5 NDCG points on the val set and by over 7 points on the test set. On the official challenge test set, MCA-I-H ranks $\nth{2}$: it improves over 7 NDCG over the best single model but loses by 2 points against a 6-strong RvA ensemble model (2019 winning entry).
 %, which gives an absolute improvement of 20 points over the LF Baseline % (with Faster-RCNN features)
%and 17 points over the SOTA RvA model (Table \ref{table:val-curriculum}).}

Compared to MCA-I, which treats the task as multiple rounds of VQA, encoding history improves results, but only significantly for MCA-I-VGH in the sparse annotation phase. After fine-tuning, MCA-I-VGH and MCA-I-H perform equally.
MCA-I-H implements a late/shallow fusion of history and image.
Architectures which model dense interaction between the conversational history and the image representations (i.e.\ MCA-I-VGH, MCA-I-HGuidedQ) perform comparably; only MCA-HConcQ performs significantly worse. 
Note that MCA-I %which considers Visual Dialog as multiple rounds of VQA (requiring no dialog history) 
also outperforms the baselines and current SOTA by a substantial margin (both in the sparse annotation phase and curriculum fine-tuning phase), while, counter-intuitively, there is not a significant boost by adding conversational history. This is surprising, considering that according to \citet{das2017visual}, 38\% of questions contain a pronoun, which would suggest that these questions would require dialog history in order to be ``understood/grounded" by the model.

Furthermore, curriculum fine-tuning  significantly improves performance with an average improvement of 11.7 NDCG points, but worsens performance in terms of the other metrics, which only consider a single ground truth (GT) answer.

\section{Error Analysis}\label{sec:erroranalysis}

In the following, we perform a detailed error analysis, investigating the benefits of dialog history encoding and the observed discrepancy between the NDCG results and the other retrieval based metrics. % to %further understand the results above. 
% In particular, we investigate the benefits of dialog history encoding and the observed discrepancy between the NDCG results and the other retrieval based methods.

\subsection{Dialog History}\label{ssec:dialoghistory}

\begin{table}[ht]
\centering
% \resizebox{0.48\textwidth}{!}{
\begin{tabular}{lrr}
\hline
\textbf{Annotation} & \textbf{Count} & \textbf{Percentage} \\
\hline \hline
VQA turns  & 594 & 67.12\% \\
\textbf{History required}  &  \textbf{97}  &  \textbf{10.96\%} \\
Common Sense  &  94 & 10.62\% \\
Guess  &  59 & 6.67\% \\
Cant tell  &  34 & 3.84\% \\
Not relevant  &  7 & 0.79\% \\
\hline
\end{tabular}
% }
\caption{Results of crowd-sourcing study to understand whether humans require dialog history to answer the question. `VQA turns' indicate that humans could potentially answer correctly without having access to the previous conversation while `History required' are the cases identified requiring dialog context. We also identified the cases requiring world knowledge/ common sense, guessing and questions not relevant to the image.}
\label{table:crowd-sourcing}
\end{table}

We performed an ablation study whereby we did not include
the caption as part of
historical context and compare with the results in Table \ref{table:val-curriculum}. The performance dropped from (NDCG 72.2, MRR 42.3)  to (NDCG 71.6, MRR 40.7) using our best performing MCA-I-H model after finetuning. Since the crowd-sourced conversation was based on the caption, the reduced performance was expected. 
% Our results show that the information provided by the caption accounted for 0.8 NDCG points and 1.6 MRR points in our best performing model.

%% SA: TODO - uncomment
In order to further verify the role of dialog history, we conduct a crowd-sourcing study to understand which questions require dialog history, in order to be understood by humans. We first test our history-encoding models on a subset (76 dialogs) of the recently released
VisPro dataset \cite{yu2019you} which focuses on the task of Visual Pronoun Resolution.\footnote{We use the intersection of dialogs in VisDial val set and VisPro to create this subset.} 
Note that VisPro also contains non-referential pleonastic pronouns, i.e.\ \ pronouns used as ``dummy subjects" when e.g.\ talking about the weather (``Is it sunny?").%\footnote{Note that questions with pleonastic pronouns such as ``does it look like\ldots" or ``it it sunny/daytime/day\ldots" are one of the most frequently asked questions according to \cite{das2017visual}. Also in appendix.}

We thus create a new crowd-sourced dataset\footnote{Data collection code available at \url{https://github.com/shubhamagarwal92/visdialconv-amt}}, which we call {\em VisDialConv}. This is a subset of the VisDial val-set %based on crowd-sourced annotations 
 consisting of 97 dialogs, where the crowd-workers identified single turns (with dense annotations) requiring historical information. In particular, we asked crowd-workers whether they could provide an answer to a question given an image, without showing them the dialog history, and select one of the categories in Table \ref{table:crowd-sourcing} (see further details in Appendix \ref{sec:amt}).
% SA: DONE - TODO - after submission
 %\ref{sec:amt}). 
 
 In order to get reliable results, we recruited 3 crowd-workers per image-question pair and only kept instances where at least 2 people agreed. Note that we only had to discharge 14.5\% of the original 1035 image-question pairs, leaving us with 885 examples. The results in Table \ref{table:crowd-sourcing} show that only 11\% required actual dialog historical context according to the crowd-workers. Most of the time (67\% cases), crowd-workers said they can answer the question correctly without requiring history. 
 
 The results in Table \ref{table:val-subset-data} are on the subset of 97 questions which the crowd-workers identified as requiring history.\footnote{We took care to only include examples from Visdial val set in both Vispro and VisDialConv subsets. Also note, there are only 8 overlapping instances between Vispro and VisdialConv subsets.} They show that history encoding models (MCA-I-HGuidedQ / MCA-I-HConcQ / MCA-I-H / MCA-I-VGH) significantly outperform MCA-I, suggesting that this data cannot be modelled as multiple rounds of VQA. It can also be seen that all the models with dense (early) interaction of the historical context outperform the one with late interaction (MCA-I-H) in terms of NDCG. %They are more 
Models with dense interactions appear to be more reliable in choosing other correct relevant answers because of the dialog context. 
%As observed previously on the whole dataset, MCA-I-VGH outperforms during the spare annotation phase. 

Our best performing model on VisDialConv is MCA-I-HGuidedQ and achieves a NDCG value of 62.9 after curriculum fine-tuning. However, on the VisPro subset, we observe that MCA-I-H still outperforms the other models. Interestingly, on this set, MCA-I also outperforms other history encoding models (except for MCA-I-H). 

In sum, our analysis shows that only a small subset of the VisDial dataset contains questions which require dialog history, and for those, models which encode history lead to better results. We posit that this is due to the fact that questions with pleonastic pronouns such as %``does it look like\ldots" or
``Is it sunny/daytime/day\ldots" are % one of 
the most frequent according to our detailed analysis in Appendix \ref{sec:top20} about the dialog phenomena. % SA - DONE - TODO - after submission
% \ref{sec:top20}.

\begin{table}[ht]
\centering
\resizebox{0.48\textwidth}{!}{
\begin{tabular}{crrrr}
\hline
\multirow{2}{*}{\textbf{Relevance of GT}} & \multicolumn{2}{c}{\textbf{Train}} & \multicolumn{2}{c}{\textbf{Val}} \\ 
\cline{2-5} 
& \textbf{Count} & \textbf{Percent} & \textbf{Count} & \textbf{Percent} \\ 
\hline \hline
1 & 1057 & 52.85\% & 643 & 31.15\% \\
0.8 & - & - & 397 & 19.23\% \\
0.6 & - & - & 330 & 15.99\% \\
0.5 & 526 & 26.30\% & - & - \\
0.4 & - & - & 281  & 13.61\% \\
0.2 & - & - & 227 & 11.00\% \\
0 & 417 & \textbf{20.85\%} & 186 & \textbf{9.01\%} \\ \hline
Total & 2000 & 100\% & 2064 & 100\% \\ 
\hline
\end{tabular}
}
\caption{Relevance score (dense annotation) provided for 2k/2k train/val QA turns. We find that 20\% of the ground truth answers were marked as irrelevant (0 score) and partially relevant (0.5 score) by the human annotators for train set. This can be attributed to human errors made while  collecting the original data as well as when crowd-sourcing the dense annotations.}
\label{table:dense}
\end{table}

%%%%%%%%%%%%%%%%%%%%%%%%%%%%%%%%%%%%%%%%%%%%%%%%%%%%%%%%%%%%%%%%%%%%%%%%%%%%%%%%%%%% Figure starts here %%%%%%%%%%%%%%%%%%%%%%%%%%%%%%%%%%%%%%%%%%

% \input{results_analysis_main.tex}
% Standalone image here
% https://www.overleaf.com/2228815533dktyqmdzxxpz

\begin{figure*}[ht]
% \begin{table*}[ht] 
\centering
\setlength{\tabcolsep}{5pt}
\resizebox{\textwidth}{!}{
\begin{tabular}{p{7cm}p{5cm}p{6.5cm}p{6.5cm}}
\hline
\textbf{\small{Image}} & \textbf{\small{Dialog}} &  \textbf{\small{MCA-I-H}} & \textbf{\small{MCA-I-VGH}} \\
\hline \hline
% Example starts here
%% Index : 45
\raisebox{-1.2in}{ 
\begin{minipage}{7cm} % bigger size decreasing effect
  \includegraphics[width=\linewidth]{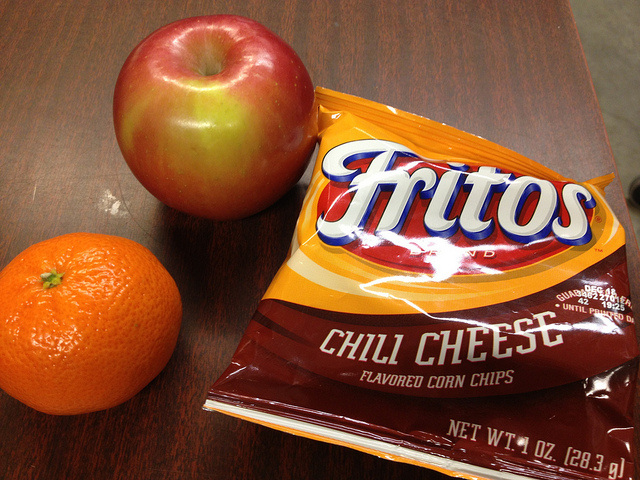}%
  \\
  A bag of chips and a apple and orange. \\
  $N_{Rel}$: 15
\end{minipage}
}
& 
% \newline
% \newline
% \newline
\tt {\textcolor{red}{Q} What kind of chips are they?
\newline
\textcolor{blue}{A} Chili cheese corn chips.
\newline
\textcolor{red}{Q} Is the bag open or still sealed?
\newline
\textcolor{blue}{A} Sealed.
\newline
\textcolor{red}{Q} Is it next to the apple and orange?
\newline
\textcolor{blue}{A} Yes.
\newline
\textcolor{red}{Q} Are they all on a table?
\newline
\textcolor{blue}{GT:} Yes.
\newline
\textcolor{blue}{Rel:} 1.0
}&

\tt{$\clubsuit$~~$R_{GT}$:1 ;  NDCG: 65.56
\begin{tabular}{cp{5cm}} %{p{2cm}p{2cm}}
(1.0) & Yes. \\
(1.0) & Yes they are on a table. \\
(0.0) & Maybe , it's a close up. \\
(0.0) & Can't see a table. \\
(0.2) & I think so, it is a close up. \\
\end{tabular}
}
\newline
\tt{$\vardiamondsuit$~~$R_{GT}$:2 ;  NDCG: 69.94
\begin{tabular}{cp{5cm}} %{p{2cm}p{2cm}}
(0.8) & I think so. \\
(1.0) & Yes. \\
(0.2) & It appears to be. \\
(0.4) & I would think so. \\
(0.2) & I think so, it is a close up. \\
\end{tabular}
}
& 

% \textbf{Spare Annotation Phase}
\tt{$\clubsuit$~~$R_{GT}$:1 ;  NDCG: 83.93 
% \vspace{2mm}
\begin{tabular}{cp{5cm}} %{p{2cm}p{2cm}}
% Rel & Ans \\
(1.0) & Yes. \\
(1.0) & Yes they are on a table. \\
(0.0) & Yes they are. \\
\\
(0.0) & Can't see a table. \\
(0.2) & I think so, it is a close up. \\
\end{tabular}
}
\newline
% \hdashline[0.5pt/2pt]
\tt{$\vardiamondsuit$~~$R_{GT}$:4 ;  NDCG: 84.15
\begin{tabular}{cp{5cm}} %{p{2cm}p{2cm}}
(0.8) & I think so. \\
(0.8) & They appear to be. \\
(0.4) & Probably. \\
(1.0) & Yes. \\
(1.0) & Yes they are. \\
\end{tabular}
}
\\
\hline
% New example
%%% Example index 26
\raisebox{-1.3in}{ 
\begin{minipage}{7cm} % bigger size decreasing effect
  \includegraphics[width=\linewidth]{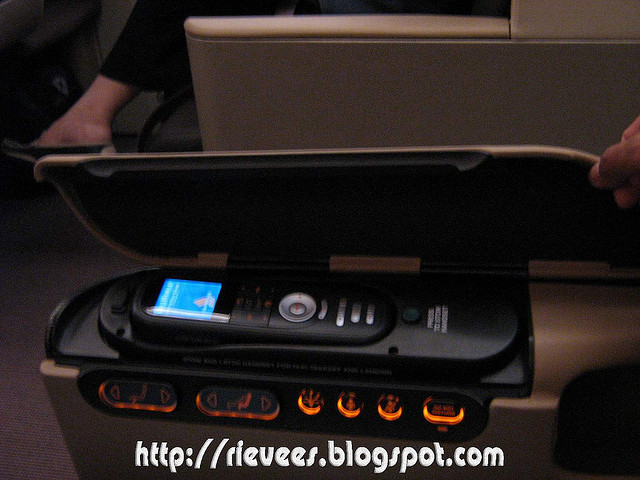}%
  \\
A remote controller is hidden in a console inside of an arm rest.\\
$N_{Rel}$: 8
\end{minipage}
}
& 
\tt {
\textcolor{red}{Q} Can you see the remote?
\newline
\textcolor{blue}{A} Yes i can.
\newline
\textcolor{red}{Q} What color is it?
\newline
\textcolor{blue}{A} It is black.
\newline
\textcolor{red}{Q} Can you tell what it is for?
\newline
\textcolor{blue}{A} It appears to be a phone.
\newline
\textcolor{red}{Q} What kind of furniture is it in?
\newline
\textcolor{blue}{GT:} Looks like a car console.
\newline
\textcolor{blue}{Rel:} 0.4
}&
\tt{$\clubsuit$~~$R_{GT}$:1 ; NDCG: 63.19
\begin{tabular}{cp{5cm}} %{p{2cm}p{2cm}}
(0.4) & Looks like a car console. \\
(0.4) & It looks like a chair on a train or a bus. \\
(0.0) & There are tables. \\
\\
(0.0) & Looks like an outdoor space. \\
(0.2) & It's a cubicle with shelves. \\
\end{tabular}
}
\tt{$\vardiamondsuit$~~$R_{GT}$:3 ; NDCG: 79.2
\begin{tabular}{cp{5cm}} %{p{2cm}p{2cm}}
(0.4) & I cannot tell. \\
(0.4) & I can't tell. \\
(0.4) & Looks like a car console. \\
(0.2) & Not sure. \\
(0.4) & Can't tell. \\
\end{tabular}
}
&
\tt{$\clubsuit$~~$R_{GT}$:2; NDCG: 58.99
\begin{tabular}{cp{5cm}} %{p{2cm}p{2cm}}
(0.0) & A cell phone, i can't see it close up. \\
(0.4) & Looks like a car console. \\
(0.4) & It looks like a chair on a train or a bus. \\
(0.2) & It's a cubicle with shelves. \\
(0.0) & The picture does not show 1. \\
\end{tabular}
}
\tt{$\vardiamondsuit$~~$R_{GT}$:7 ; NDCG: 82.22
\begin{tabular}{cp{5cm}} %{p{2cm}p{2cm}}
(0.4) & I cannot tell. \\
(0.4) & Can't tell. \\
(0.4) & I can't tell. \\
\\
(0.2) & Not sure. \\
(0.0) & The picture does not show 1. \\
\end{tabular}
}
\\
\hline
\end{tabular}
}
\caption{Top-5 ranked predictions (relevance in parentheses) of MCA-I-H and MCA-I-VGH after both sparse annotation and curriculum fine-tuning phase. $R_{GT}$ defines the rank of Ground Truth (GT) predicted by the model. We also calculate NDCG of rankings for current question turn. $N_{Rel}$ denotes number of candidate answer options (out of 100) with non-zero relevance (dense annotations). Here $\clubsuit$ and $\vardiamondsuit$ represents predictions after sparse annotation and curriculum fine-tuning respectively.} %More examples in appendix.} %\VR{Why do we need more examples?}}
\label{fig:analysis}
% \end{table*}
\end{figure*}

%%%%%%%%%%%%%%%%%%%%%%%%%%%%%%%%%%%%%%%%%%

\subsection{Dense Annotations for NDCG}
\label{subsec:ndcg}

Here, we investigate the discrepancy between the NDCG results and the other retrieval-based methods. First, we find that the annotation scales differs: while there is a 3-way annotation on the train set, the val set defines 6 possible relevance classes, see Table \ref{table:dense}. This affects the evaluation results of our model, for which we can't do much.

Next, a manual inspection reveals that the relevance weight annotations contain substantial noise: We find that ground truth answers were marked as irrelevant for about 20\% of train and 10\% of val set. Thus, our models seem to get ``confused" by fine-tuning on this data. We, therefore, manually corrected the relevance of only these GT answers (in dense annotations of train set only, but not in val set). Please see Appendix \ref{sec:dense-annotations} 
% SA - DONE - TODO - after submission
%\ref{sec:dense-annotations} 
for further details. The results in Table \ref{table:val-curriculum} (for MCA-I-H-GT) show that the model fine-tuned on the corrected data still achieves a comparable NDCG result, but substantially improves stricter (single answer) metrics, which confirms our hypothesis.
% {\bf MCA-I-H-GT in Table \ref{table:val-curriculum}}

Finally, due to the noisy signal they receive during fine-tuning, our models learn to select ``safe" answers\footnote{We show the statistics of top-ranked predictions by our MCA-I-H model on our VisdialConv subset (i.e. 97 dialogs of the Visdial val set).  Read as: (Response, count, \%) (Yes, 14, 14\%) (No, 11,  11.34\%) (I cannot tell, 9, 9.27\%) (Nope, 3, 3\%) (Not that I see, 2, 2.06\%) (Red and white, 2, 2.06\%) (Not sure, 2, 2.06\%) (I can't tell, 2, 2.06\%). This shows that at least 13.3\% of answers are non-commital (I cannot tell, Not sure, I can’t tell).}, such as ``I can't tell" (see examples in Figure \ref{fig:analysis}), which rank high according to (the more forgiving) NDCG, but perform poorly for stricter metrics like MRR and Recall. %In fact, results for R@5 and R@10 are less affected after fine-tuning, confirming our hypothesis.

%%% TODO - SA: Check if it is table or figure

\section{Discussion and Related Work}
\label{sec:related work}

Our results suggest that the VisDial dataset only contains very limited examples which require dialog history. Other visual dialog tasks, such as  GuessWhich? \cite{chattopadhyay2017evaluating} and GuessWhat?! \cite{de2017guesswhat} take place in a goal-oriented setting, which according to \newcite{schlangen2019grounded}, will lead to data containing more natural dialog phenomena. However, there is very limited evidence that  dialog history indeed matters for these tasks \cite{historyGuessWhat:2019}.
As such, we see data collection to capture visual dialog phenomena as an open problem. %To support this claim we have provided an additional analysis of dialog phenomena, including pronouns and ellipsis in Appendix E. 
% SA - TODO - after submission
% \ref{sec:heuristics}.

Nevertheless, our results also show that encoding dialog history still leads to improved results. This is in contrast with early findings that a) ``naive" encoding will harm performance (\citet{das2017visual}; see MCA-I-HConcQ in Table \ref{table:val-curriculum}), or that b) history is not necessary \citep{massiceti2018visual}. %; see OnlyQA vs MCA-H in Table \ref{table:val-curriculum}).

Furthermore, we find that our model learns to provide generic answers by taking advantage of the NDCG evaluation metric. Learning generic answers is a well-known problem for open-domain dialog systems, e.g.\ \cite{li-etal-2016-diversity}. While the dialog community approaches these phenomena by e.g. learning better models of coherence \cite{xu-etal-2018-better}, we believe that evaluation metrics also need to be improved for this task, as widely discussed for other generation tasks, e.g.\ \cite{liu2016not, novikova2017we, reiter2018structured}. As a first step, BERT score~\cite{zhang2019bertscore} could be explored to measure ground-truth similarity replacing the noisy NDCG annotations of semantic equivalence.

%%%%%%%%%%%%%%%%%%%%%%%%%%%%%%%%%%%%%%%%%%
\section{Conclusion and Future Work}
\label{sec:conclusion}

In sum, this paper shows that we can get SOTA performance on the VisDial task by using transformer-based models with Guided-Attention \cite{yu2019deep}, and by encoding dialog history and fine-tuning we can improve results even more. 

Of course, we expect %models using 
pre-trained visual BERT models to show even more %significant
improvements on this task, e.g. %VL-bert \cite{su2019vl}, 
Vilbert \cite{lu2019vilbert}, LXMert \cite{tan2019lxmert}, UNITER \cite{chen2019uniter} etc. However, we also show the limitations of this shared task in terms of dialog phenomena and evaluation metrics. We, thus, argue that progress needs to be carefully measured by posing the right task in terms of dataset and evaluation procedure.

% TODO - uncomment this later
\section*{Acknowledgments}
We thank the anonymous reviewers for their insightful comments. Shubham would like to thank Raghav Goyal for the discussions during `Pikabot' submission to Visual Dialog Challenge 2018. This work received continued support by Adobe Research gift funding for further collaboration. This research also received funding from Adeptmind Inc., Toronto, Canada and the EPSRC project MaDrIgAL (EP/N017536/1). We would also like to acknowledge the AWS Cloud Credits for Research programme.

\bibliography{visdial}
\bibliographystyle{acl_natbib}

% TODO - uncomment this later
\appendix
\section{More implementation details}
%%%%%%%%%%%%%%%%%%%%%%%%%%%%%%%%%%%%%%%%%%%%
\label{sec:more-implementation}

We built our implementation upon
starter code in PyTorch which the VisDial organisers kindly provided.\footnote{\url{https://github.com/batra-mlp-lab/visdial-challenge-starter-pytorch}.} We follow the guidelines of \newcite{teney2018tips} and used static 36 as the number of object proposals in our experiments (though our model can handle dynamic number of proposals). 
% We experimented with different learning rates and used 

We experimentally determined the learning rates of 0.0005 for training MCA models and 0.0001 for fine-tuning %. LR during training was reduced
  and reducing it by 1/10 after every 7 and 10 epochs out of a total of %and we train the model for 
  12 epochs for training and 1/5 after 2 epochs for fine-tuning.
  
    We use pytorch's LambdaLR scheduler while training and ReduceLROnPlateau for the fine-tuning procedure. Dropout of 0.2 is used for regularization and we perform early stopping and saved the best model by tracking the NDCG value on val set. Layer normalisation ~\cite{ba2016layer} is used for stable training following \cite{vaswani2017attention,yu2019deep}. Attention reduction consisted of 2 layer MLP (fc(d)-ReLU-Dropout(0.2)-fc(1)).

We also experimented with different contextual representations, including BERT \cite{devlin2019bert}; However we didn't observe any improvement, similar to the observation by \cite{tan2019lxmert}.

For the results on the validation set, only the training split is used. To report results on test-std set, both the training and val set are used for training. For curriculum fine-tuning we use multi-class cross entropy loss where weighted by the relevance score. %\footnote{Note: this is different from weights provided in nn.CrossEntropyLoss which weighs the class for imbalanced dataset.}
All our MCA modules have 6 layers and 8 heads, which we determined via a hyper parameter search. 
%We also conducted a small hyper parameter search of the number of layers and number of heads in the MCA module.
Table \ref{table:hyperparameter} shows more details. 

%% Table with hyper-parameters

\begin{table}[ht!]
\centering
\resizebox{0.48\textwidth}{!}{
\begin{tabular}{lp{75mm}}
\hline
\textbf{Annotation} & \textbf{Text} \\
\hline \hline
VQA turns  & I can confidently tell the correct answer just seeing the image. \\
\textbf{History required}  &  I want to know what was discussed before to answer confidently. Cannot answer with just the question and image. Need more information (context) from previous conversation. \\
Common Sense  &  I can answer it but by inferring using common sense. \\
Guess  &  I can only guess the answer. \\
Cant tell  & I can't tell the answer. \\ 
Not relevant  &  Not relevant question for this image. \\
\hline
\end{tabular}
}
\caption{Mapping of human annotation with the actual text shown to the user.
}
\label{table:crowd-sourcing-annotation}
\end{table}

\begin{table*}[ht!]
\centering
\resizebox{\textwidth}{!}{
\begin{tabular}{l|cccccc|cccccc}
\hline
\multicolumn{1}{l|}{\multirow{2}{*}{\textbf{Model}}} & \multicolumn{6}{|c|}{\textbf{Training}} & \multicolumn{6}{|c}{\textbf{Curriculum Fine-tuning}} \\ \cline{2-13} 
\multicolumn{1}{l|}{}                       & \multicolumn{1}{c}{\textbf{NDCG}} & \multicolumn{1}{c}{\textbf{MRR}} & \multicolumn{1}{c}{\textbf{R@1}} & \multicolumn{1}{c}{\textbf{R@5}} & \multicolumn{1}{c}{\textbf{R@10}} & \multicolumn{1}{c|}{\textbf{Mean}} & \multicolumn{1}{c}{\textbf{NDCG}} & \multicolumn{1}{c}{\textbf{MRR}} & \multicolumn{1}{c}{\textbf{R@1}} & \multicolumn{1}{c}{\textbf{R@5}} & \multicolumn{1}{c}{\textbf{R@10}} & \multicolumn{1}{c}{\textbf{Mean}} \\ 
\hline \hline

MCA-I-H (L6 H8) & 60.27 & 64.33 & 51.12 & 80.91 & 89.65 & 4.24 & 72.22 & 42.38 & 26.94 & 60.17 & 75.2 & 8.2 \\
MCA-I-H (L2 H4) & 58.99 & 64.46 & 51.14 & 81.03 & 89.91 & 4.19 & 70.57 & 42.48 & 26.3 & 61.3 & 76.05 & 8.06 \\
MCA-I-H (L6 H2) & 60.13 & 60.63 & 46.7 & 77.55 & 87.47 & 4.8 & 70.42 & 39.17 & 23.3 & 57.64 & 73.48 & 8.69 \\
\hline
\end{tabular}
}
\caption{Hyper-parameter tuning for  number of layers and number of heads. The results in the main paper are reported with 6 Layers(L6) and 8 Heads (H8) for all MCA models.}
\label{table:hyperparameter}
\end{table*}

\section{AMT Interface}
\label{sec:amt}

\begin{figure*}[ht!]
\centering
\includegraphics[scale=0.4]{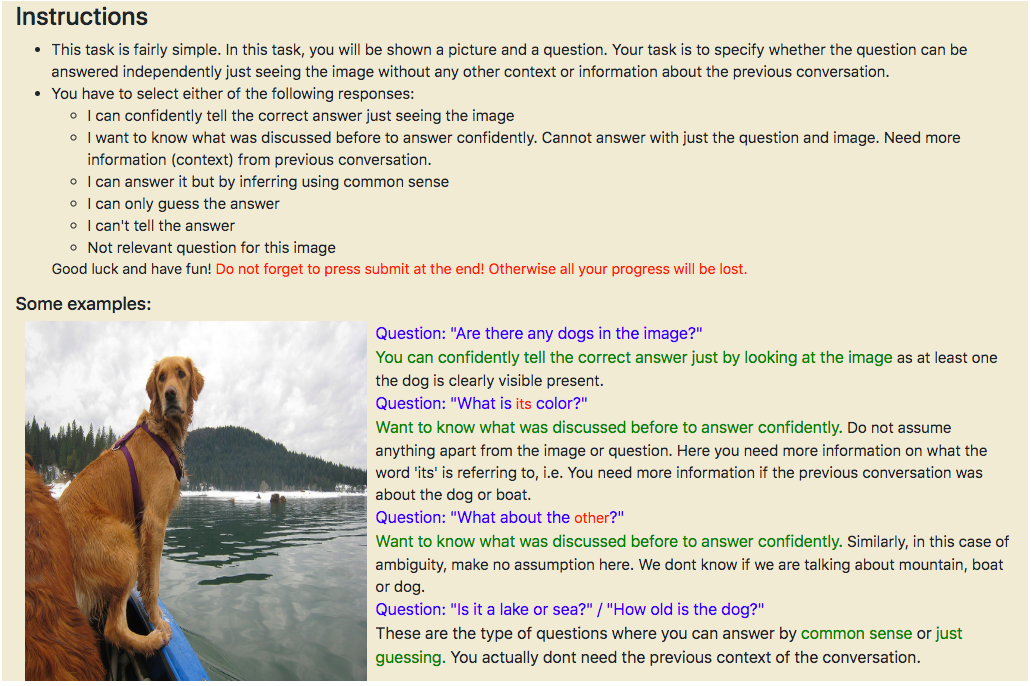}
\caption{Instructions for the AMT task.}
\label{fig:instructions}
\end{figure*}

\begin{figure*}[ht!]
\centering
\includegraphics[scale=0.48]{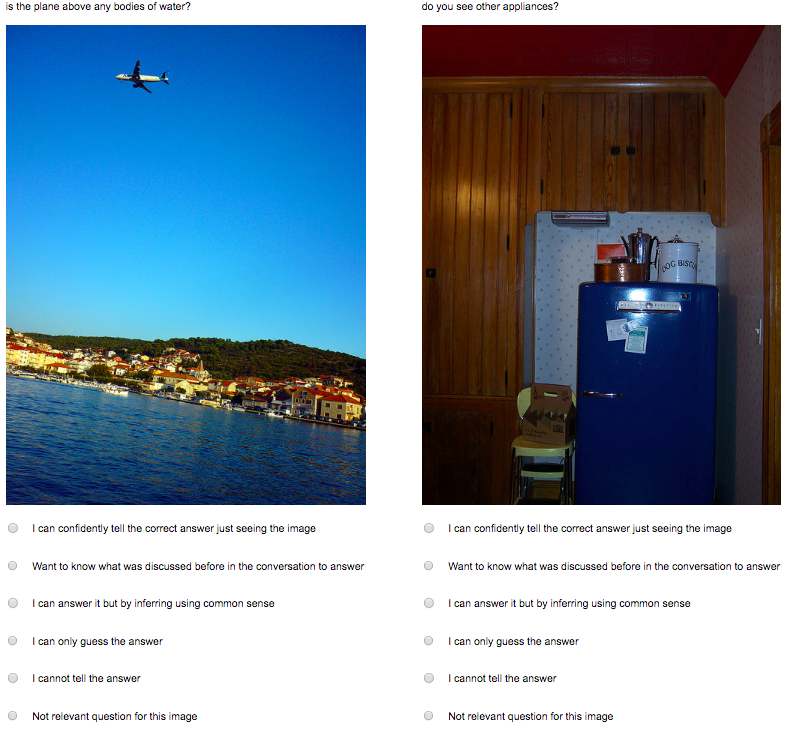}
\caption{Qualification test consisting of 2 test images to allow the turkers to actually attempt the task}
\label{fig:qual-test}
\end{figure*}

\begin{figure*}[ht!]
\centering
\includegraphics[scale=0.4]{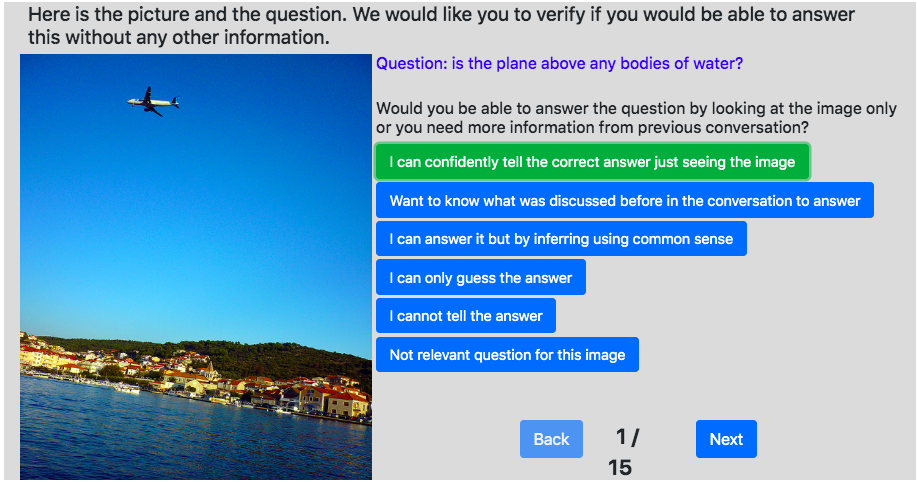}
\caption{Sample task.}
\label{fig:amt-task}
\end{figure*}

Here, we provide more details on the crowd-sourcing study described in Section \ref{ssec:dialoghistory}.
%% TODO - SA - After submission - convert it back 
% .
Figure ~\ref{fig:instructions} shows the instructions shown to the turkers. We also setup a qualification test consisting of 2 test images (in Figure \ref{fig:qual-test}) to assess whether turkers understood the task properly. 
This allowed us to have an automated quality check for the annotations. Each HIT consisted of 15 images. For the actual task (e.g. Fig. \ref{fig:amt-task}), users were shown just the image and the current question -- without any previous historical context -- and asked to choose one of the answers as shown in Table \ref{table:crowd-sourcing-annotation}. Our AMT interface\footnote{We built upon the repo: \url{https://github.com/jcjohnson/simple-amt}.} used AWS boto3 library in python. %Code released after acceptance.

\begin{figure*}[ht!]
% \begin{table*}[ht]
\centering
\setlength{\tabcolsep}{5pt}
\resizebox{\textwidth}{!}{
\begin{tabular}{p{7cm}p{5cm}p{6.5cm}p{6.5cm}}
\hline
\textbf{\small{Image}} & \textbf{\small{Dialog}} &  \textbf{\small{MCA-I-H}} & \textbf{\small{MCA-I-VGH}} \\
\hline \hline
% Example starts here
%% Index : 45
\raisebox{-1.13in}{ 
\begin{minipage}{7cm} % bigger size decreasing effect
  \includegraphics[width=\linewidth]{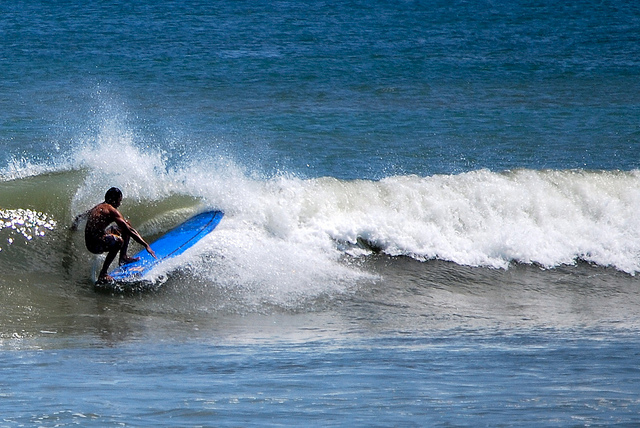}%
  \\
    A surfer crouches as they ride a cresting wave.\\
    $N_{Rel}$: 15
\end{minipage}
}
& 
% \newline
\tt{\textcolor{red}{Q} Is the photo in color?
\newline
\textcolor{blue}{A} Yes.
\newline
\textcolor{red}{Q} Any other people?
\newline
\textcolor{blue}{GT:} No.
\newline
\textcolor{blue}{Rel:} 0.8
}&

\tt{$\clubsuit$~~$R_{GT}$:1 ; NDCG 83.32
\begin{tabular}{cp{5cm}} %{p{2cm}p{2cm}}
(0.8) & No. \\
(0.2) & 0. \\
(1.0) & Nope. \\
(0.8) & No there's not. \\
(0.4) & Just the 1. \\
\end{tabular}
}
\newline
\tt{$\vardiamondsuit$~~$R_{GT}$:2; NDCG 91.2 
% \newline
\begin{tabular}{cp{5cm}} %{p{2cm}p{2cm}}
% Rel & Ans \\
(1.0) & Nope. \\
(0.8) & No. \\
(0.8) & Not that i can see. \\
(0.8) & Not that i see. \\
(0.8) & No there's not. \\
\end{tabular}
}
& 
\tt{$\clubsuit$~~$R_{GT}$:1 ;  NDCG 74.98
\begin{tabular}{cp{5cm}} %{p{2cm}p{2cm}}
(0.8) & No. \\
(0.2) & 0. \\
(1.0) & Nope. \\
(0.2) & 0 at all. \\
(0.8) & Not that i can see. \\
\end{tabular}
}
\newline
\tt{$\vardiamondsuit$~~$R_{GT}$:1 ;  NDCG 85.24
\begin{tabular}{cp{5cm}} %{p{2cm}p{2cm}}
(0.8) & No. \\
(0.8) & Not that i can see. \\
(0.8) & Not that i see. \\
(0.8) & No there's not. \\
(1.0) & Nope. \\
\end{tabular}
}
\\
\hline
% New example
% New example
%%% Example index 26
\raisebox{-1.3in}{ 
\begin{minipage}{7cm} % bigger size decreasing effect
  \includegraphics[width=\linewidth]{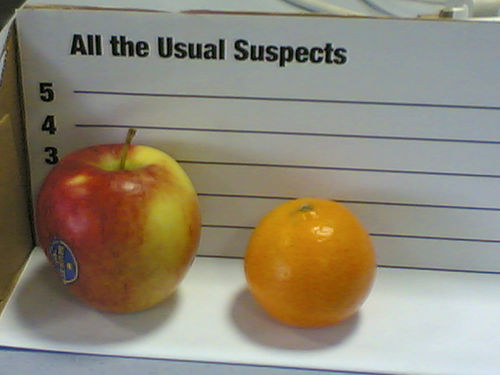}%
  \\
An apple and orange are sitting in a white box with size measurements.\\
$N_{Rel}$: 4
\end{minipage}
}
& 
\tt{
\textcolor{red}{Q} What color is the apple?
\newline
\textcolor{blue}{A} It is red and yellow.
\newline
\textcolor{red}{Q} What color is the orange?
\newline
\textcolor{blue}{A} It is dark orange.
\newline
\textcolor{red}{Q} What is the size of the box?
\newline
\textcolor{blue}{A} Can't tell.
\newline
\textcolor{red}{Q} Where is the box?
\newline
\textcolor{blue}{GT:} Can't tell.
\newline
\textcolor{blue}{Rel:} 1.0
}&
\tt{$\clubsuit$~~$R_{GT}$:9 ;  NDCG 0.0
\begin{tabular}{cp{5cm}} %{p{2cm}p{2cm}}
% Rel & Ans \\
(0.0) & On the table. \\
(0.0) & In a container by the window. \\
(0.0) & On table. \\
\\
(0.0) & Use no paper here. \\
(1.0) & I cannot tell. \\
\end{tabular}
}
\tt{$\vardiamondsuit$~~$R_{GT}$:5 ;  NDCG 73.59
\begin{tabular}{cp{5cm}} %{p{2cm}p{2cm}}
% Rel & Ans \\
(1.0) & I cannot tell. \\
(0.0) & On the table. \\
(1.0) & I can't tell. \\
(0.6) & Not sure. \\
(1.0) & Can't tell. \\
\end{tabular}
}
&
\tt{$\clubsuit$~~$R_{GT}$:9 ;  NDCG 0.0
\begin{tabular}{cp{5cm}} %{p{2cm}p{2cm}}
% Rel & Ans \\
(0.0) & On table. \\
(0.0) & On the table. \\
\\
(0.0) & In a container by the window. \\
(0.0) & In a kitchen. \\
(1.0) & I cannot tell. \\
\end{tabular}
}
\tt{$\vardiamondsuit$~~$R_{GT}$:4 ;  NDCG 98.84
\begin{tabular}{cp{5cm}} %{p{2cm}p{2cm}}
% Rel & Ans \\
(1.0) & I cannot tell. \\
(1.0) & I can't tell. \\
(0.6) & Not sure. \\
(1.0) & Can't tell. \\
(0.0) & In a container by the window. \\
\end{tabular}
}
\\
\hline
%%% Example index 26
\raisebox{-1.3in}{ 
\begin{minipage}{7cm} % bigger size decreasing effect
  \includegraphics[width=\linewidth]{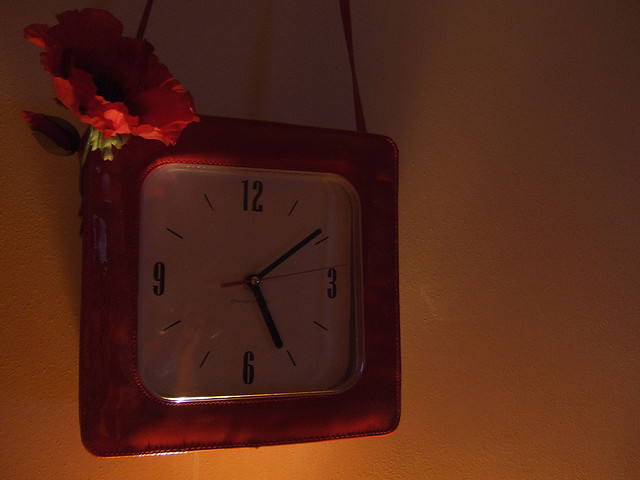}%
  \\
A clock with a rose on its corner sits on the wall.\\
$N_{Rel}$: 3
\end{minipage}
}
& 
\tt {
\textcolor{red}{Q} Is it a real rose?
\newline
\textcolor{blue}{A} I don't think so.
\newline
\textcolor{red}{Q} Is it an old fashion clock?
\newline
\textcolor{blue}{A} No, not really.
\newline
\textcolor{red}{Q} Is it a digital clock?
\newline
\textcolor{blue}{A} No.
\newline
\textcolor{red}{Q} Is it hanging on the wall or leaning?
\newline 
\textcolor{blue}{GT:} It's hanging.
\newline
\textcolor{blue}{Rel:} 1.0
}&
\tt{$\clubsuit$~~$R_{GT}$:1 ;  NDCG: 81.55
\begin{tabular}{cp{5cm}} %{p{2cm}p{2cm}}
(1.0) & It's hanging. \\
\\
(0.0) & Yes, it's attached to the side of the building. \\
(0.0) & Yes. \\
\\
\\
(0.0) & It is cut out, but it is definitely sitting on something. \\
(0.0) & It looks like. \\
\end{tabular}
}
\tt{$\vardiamondsuit$~~$R_{GT}$:2 ;  NDCG 51.45 
\begin{tabular}{cp{5cm}} %{p{2cm}p{2cm}}
(0.0) & It looks like. \\
\\
(1.0) & It's hanging. \\
(0.0) & Can't tell. \\
(0.0) & Unclear. \\
(0.0) & I think so. \\
\end{tabular}
}
&
\tt{$\clubsuit$~~$R_{GT}$:2 ;  NDCG 51.45 
\begin{tabular}{cp{5cm}} %{p{2cm}p{2cm}}
(0.0) & No it is not mounted on the wall. \\
(1.0) & It's hanging. \\
\\
\\
(0.0) & It is cut out, but it is definitely sitting on something. \\
(0.0) & Yes, it's attached to the side of the building. \\
(0.0) & On the rail. \\
\end{tabular}
}
\tt{$\vardiamondsuit$~~$R_{GT}$:3 ;  NDCG 40.78
\begin{tabular}{cp{5cm}} %{p{2cm}p{2cm}}
(0.0) & No it is not mounted on the wall. \\
(0.0) & Not sure. \\
(1.0) & It's hanging. \\
(0.0) & Can't tell. \\
(0.0) & I can't tell. 
\end{tabular}
}
\\
\hline
\end{tabular}
}
\caption{Top-5 ranked predictions (relevance in parentheses) of MCA-I-H and MCA-I-VGH after both sparse annotation and curriculum fine-tuning phase. $R_{GT}$ defines the rank of Ground Truth (GT) predicted by the model and NDCG of rankings for current question turn. $N_{Rel}$ denotes number of candidate answer options (out of 100) with non-zero relevance (dense annotations). Here $\clubsuit$ and $\vardiamondsuit$ represents predictions after sparse annotation and curriculum fine-tuning respectively.}
\label{fig:analysis-supp}
% \end{table*}
\end{figure*}

% \begin{figure}[ht]
% \end{figure}

% \begin{figure}[ht]
% \end{figure}

% \begin{figure}[ht]
% \end{figure}

% \begin{figure}[ht]
\begin{figure*}[ht!]
\centering
\includegraphics[scale=0.45]{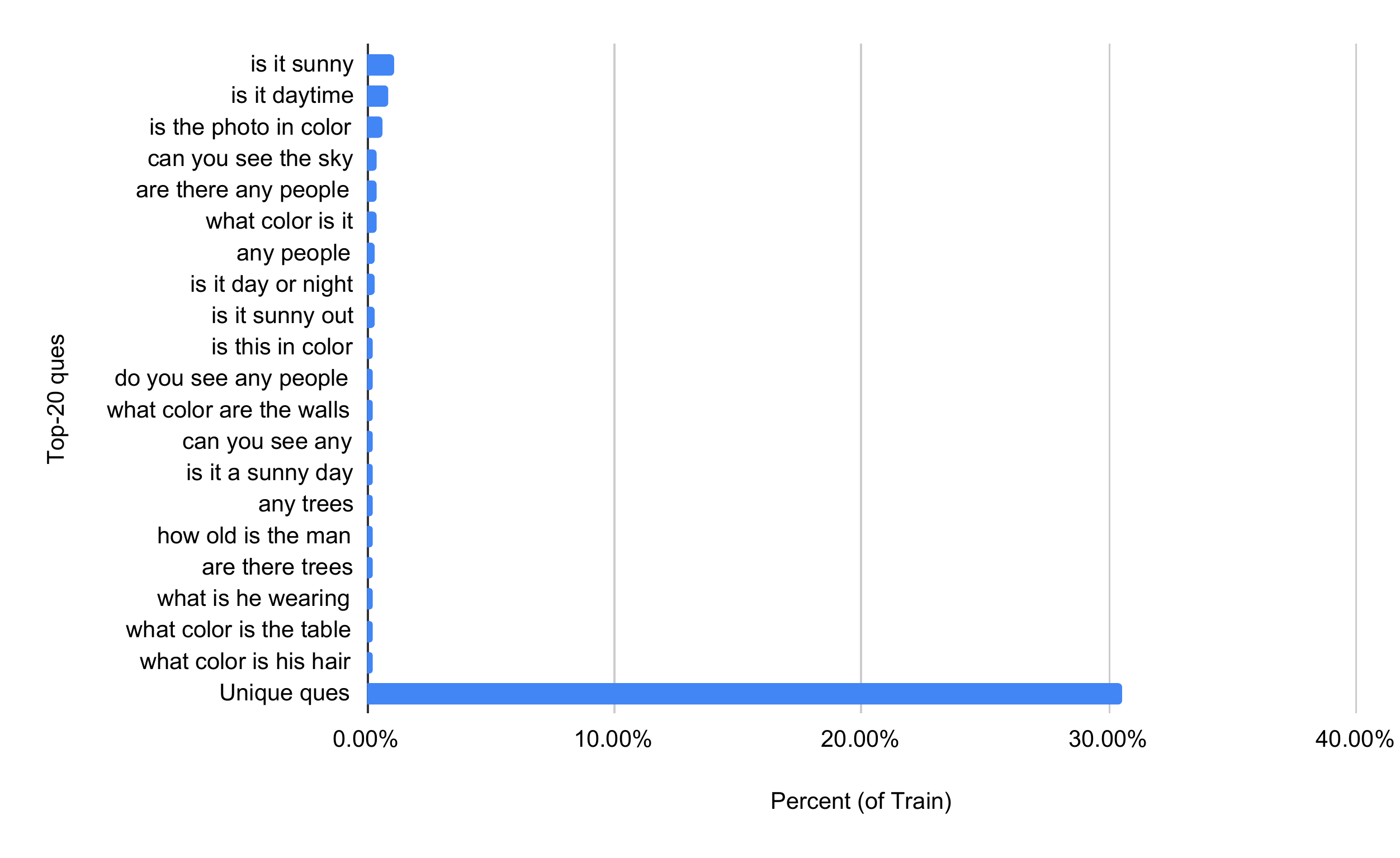}
\caption{Top-20 questions in the training set. Of all the questions in the training set, only 30\% questions are unique while weather related questions (like sunny, daytime, rainy) top the charts. }
\label{fig:top20q}
\end{figure*}
% \end{figure}

% \begin{figure}[ht]
\begin{figure*}[ht!]
\centering
\includegraphics[scale=0.44]{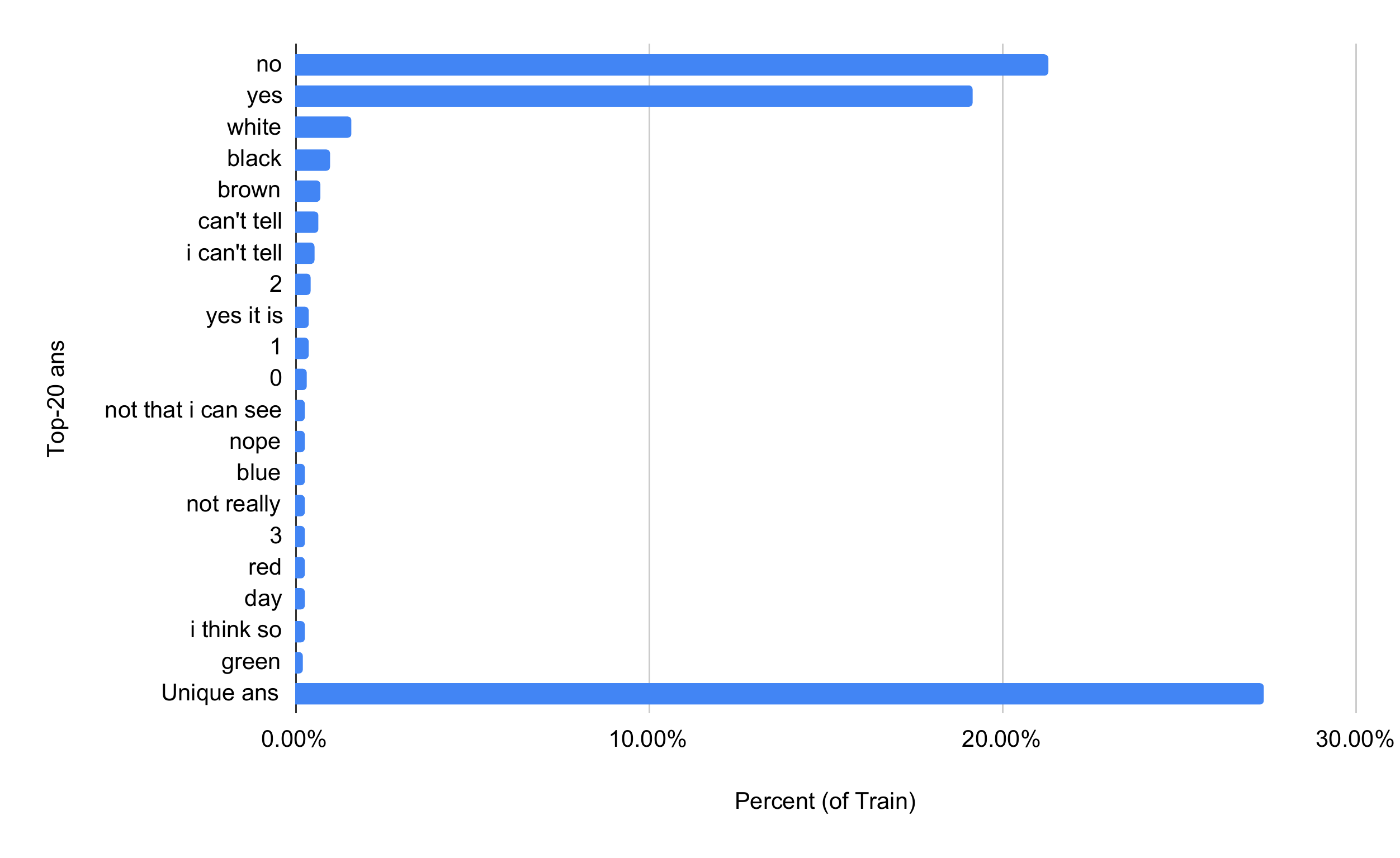}
\caption{Top-20 answers in the training set. Yes/No forms a major chunk in top 20 answers.}
\label{fig:top20a}
\end{figure*}
% \end{figure}

% \begin{figure}[ht]
\begin{figure*}[ht!]
\centering
\includegraphics[scale=0.6]{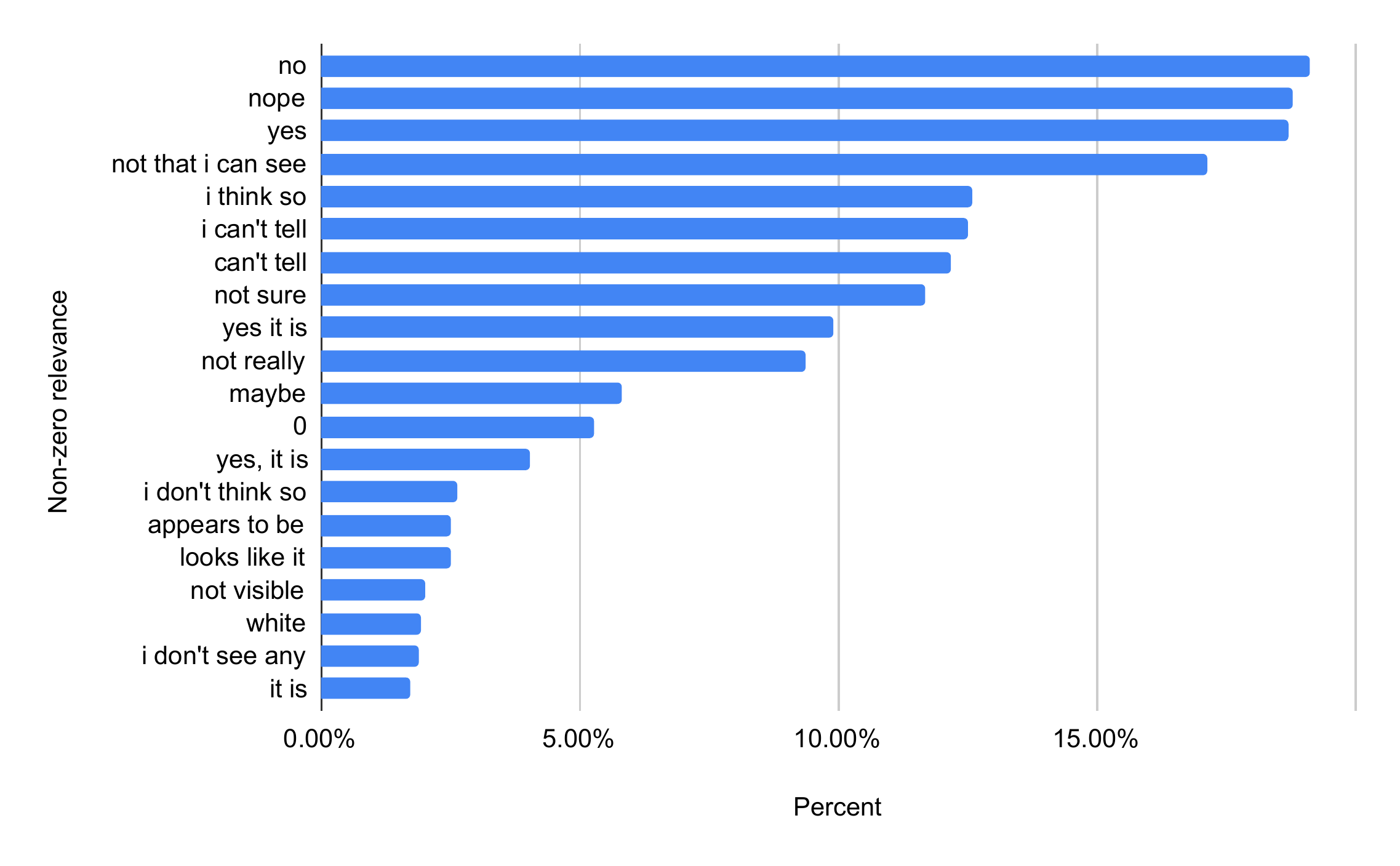}
\caption{Top-20 answers with non-zero relevance in the dense annotations of training set. Generic and yes/no semantically equivalent answers mostly constitute the list. Percentage is calculated out of total 3652 unique answers which have non-zero relevance in train dense annotations set.}
\label{fig:top20a-dense}
\end{figure*}
% \end{figure}

\section{Diversity and dialog phenomena in VisDial dataset} %: Top-20 questions, answers and candidate options}
\label{sec:top20}
% \VR{Are you using this argument somehow?}
We also did an analysis of the top-20 questions (Figure \ref{fig:top20q}) and answers (Figure \ref{fig:top20a}) in the training set. `Yes'/`No' binary answers form the major chunk (19.15\% and 21.2\% respectively) of ground truth answers. Color related answers (such as White, Brown in the top-20 answers) form 4\% of all the ground truth answers. Numbered answers (such as 0, 1, 2 ,3) form 1.3\% while `Can't tell' form another 1.2\%. 

As evident in the top-20 questions, weather related questions (such as `Is it sunny/daytime/day/night?'), color related (`What color is it/his hair/the table?') and basic conversational-starters  (`Can you see any people?') form the major portion. 

We also tried to analyze the top-20 answers (Figure \ref{fig:top20a-dense}) which had non-zero relevance in the dense annotations. Specifically, we took all 2k example turns of training set with dense annotations for each of 100 options. We find that generic answers such as `Can't tell', binary answers `Yes/No' and their semantically equivalent answers `Not that i can see' are mostly given non-zero relevance by crowd-workers.

We tried to calculate the statistics of the pronouns and ellipsis which we consider essential (but not complete) phenomena in a dialog dataset. Figure \ref{fig:pronouns} shows the number of pronouns in a dialog. We find that major chunk consisted of 2-6 pronouns in all the 10 questions across the dialog. We tried to distinguish between the usage of `it' as pleonastic and non-pleonastic pronouns (discussed in \cite{loaiciga2017disambiguating}). For e.g.\ in the sentence: `It is raining'. Here, though, `it' would be identified as a pronoun, but it doesn't refer to anything. Notice the drift in distribution of the number of pronouns (All pronouns vs Non-pleonastic). We also tried to identify the cases of ellipsis (methodology explained further) and found that majority questions (82\%) doesn't contain any case of ellipsis in the dialog. We define simple heuristics to identify dialog phenomena. Specifically, our heuristics can be listed as: 

% https://github.com/allenai/allennlp/blob/master/allennlp/pretrained.py#L130
\begin{itemize}
    \item We use constituency parser ~\cite{joshi2018extending} to parse each question\footnote{\url{https://github.com/allenai/allennlp/blob/master/allennlp/pretrained.py}}. If the parsed tree doesn't contain `Sentence' as the root (`S', `SQ', `SBARQ', `SINV'), we consider it a case of ellipsis. 
    \item We use spaCy \footnote{\url{https://spacy.io/usage/linguistic-features}} to extract the pronouns in all the questions of a dialog. 
    \item To distinguish between different usage of `it', we mark all the co-occurrences of manually defined weather identifiers (`rainy', `sunny', `daytime', `day', `night') as pleonastic.
    \item Though `other' is a pronoun, it is not tagged by standard taggers. We explicitly deal with these cases to tag `other' as a case of pronoun. For e.g. `What about the other?' 
\end{itemize}

% \begin{figure}[ht!]
% \centering
% \includegraphics[scale=0.42]{pics/dialog_phenomena/ellipsis.pdf}
% \caption{Number of cases of ellipsis in 10 questions of a dialog}
% \label{fig:ellipsis}
% \end{figure}

\section{Corrected dense annotations}
% \section{Other strategies for training and fine-tuning using dense annotations}
\label{sec:dense-annotations}

We maintain the whole relevance list, however we change the relevance of only the ground truth (GT) to 1 instead of 0/0.5 in the train annotations (only 943 values). This was done to avoid extra gradient information that the model will receive because of noise in the dataset, since these examples were already seen during the spare annotation phase. Val annotations remains unaffected for fair comparison. As expected, this simple correction increase the ground truth related metrics such as R\{1,5,10\} drastically.

\begin{figure*}[ht!]
\centering
\includegraphics[scale=0.55]{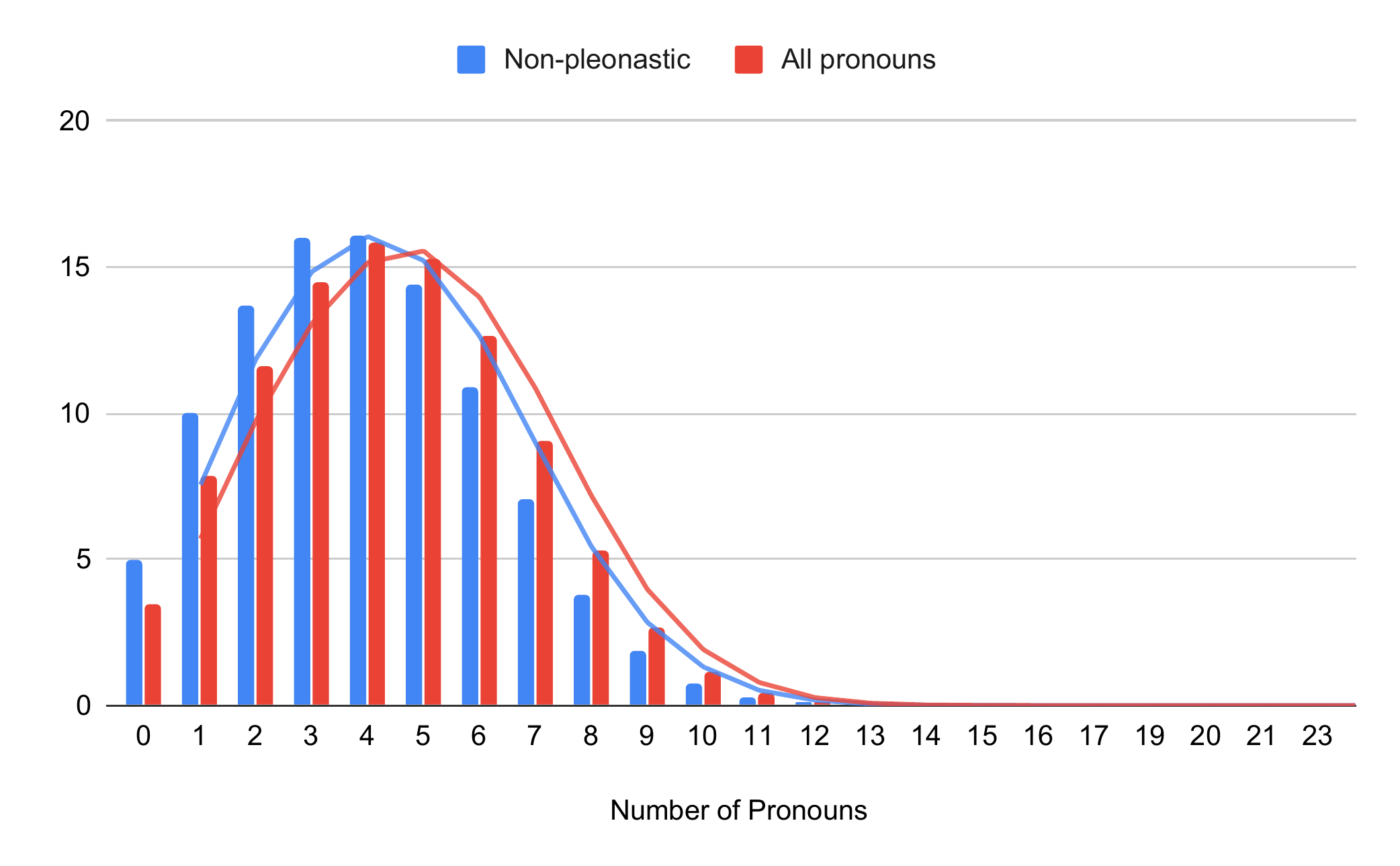}
\caption{Number of pronouns in 10 questions of a dialog.}
\label{fig:pronouns}
\end{figure*}

\ignore{
\section{Analysis and Heuristics for automatic search of dialog phenomena}
\label{sec:heuristics}

We tried to calculate the statistics of the pronouns and ellipsis which we consider essential (but not complete) phenomena in a dialog dataset. Figure \ref{fig:pronouns} shows the number of pronouns in a dialog. We find that major chunk consisted of 2-6 pronouns in all the 10 questions across the dialog. We tried to distinguish between the usage of `it' as pleonastic and non-pleonastic pronouns (discussed in \cite{loaiciga2017disambiguating}). For eg in the sentence:`It is raining'. Here though `it' would be identified as pronoun but doesn't refer to anything. Notice the drift in distribution of the number of pronouns (All pronouns vs Non-pleonastic). We also tried to identify the cases of ellipsis (methodology explained further) and found that majority questions (82\%) doesn't contain any case of ellipsis in the dialog. 

We defined simple heuristics to automatically extract cases with dialog phenomena. Specifically, our heuristics can be defined as: 

% https://github.com/allenai/allennlp/blob/master/allennlp/pretrained.py#L130
\begin{itemize}
    \item We used constituency parser ~\cite{joshi2018extending} \footnote{\url{https://github.com/allenai/allennlp/blob/master/allennlp/pretrained.py}} to parse each question. If the parsed tree doesn't contain `Sentence' as the root (`S', `SQ', `SBARQ', `SINV'), we consider it a case of ellipsis. 
    \item We used spaCy \footnote{\url{https://spacy.io/usage/linguistic-features}} to extract all the pronouns in all the questions of a dialog. 
    \item To distinguish between different usage of `it', we mark all the co-occurrences of manually defined weather identifiers (`rainy', `sunny', `daytime', `day', `night') as pleonastic.
    \item Though `other' is a pronoun it is not tagged as pronoun by standard taggers. We explicitly deal with these cases to tag `other' as a case of pronoun. For e.g. `what about the other' 
\end{itemize}

Table \ref{table:heuristics} provides the Precision-Recall (PR) metrics of our heuristics validated on the crowd-sourced VisDialConv dataset which the human annotators agreed that they required historical context and can be considered as ground truth examples which exhibit dialog phenomena. Specifically, to calculate these metrics (showing for pronouns, extended analogously to other heuristics), we consider:
\begin{itemize}
    \item True Positive (TP): Cases with pronoun and marked \textbf{requiring} historical information
    \item False Positive (FP): Cases with pronoun and marked \textbf{not requiring} historical information
    \item False Negative (FN): Cases with \textbf{no} pronoun and marked requiring historical information
    \item True Negative (TN): Cases with \textbf{no} pronoun and marked \textbf{not requiring} historical information
\end{itemize}

% In future, we plan to refine our heuristic to automate the process of subsetting the examples which exhibit dialog phenomena.

\begin{table}[ht!]
\centering
% \resizebox{\textwidth}{!}{
\begin{tabular}{ll}
% \hline
% \multicolumn{1}{l}{\textbf{VisPro subset dataset}}  \\
\hline
Metric & Percentage \\
\hline \hline
\textbf{Non-pleonastic pronoun} \\
\hline
Precision & 19.63 \\
Recall & 65.98 \\
F-score & 30.26 \\
\hline
\textbf{Pronouns} \\
\hline
Precision & 17.62 \\
Recall & 67.01 \\
F-score & 27.9 \\
\hline
\textbf{Pronouns} \\
\hline
Precision & 27.12 \\
Recall & 16.49 \\
F-score & 20.51 \\
\hline

\end{tabular}
% }
\caption{PR metrics of our heuristics on the crowd-sourced VisDialConv turns.}
\label{table:heuristics}
\end{table}

\begin{figure}[ht!]
\centering
\includegraphics[scale=0.42]{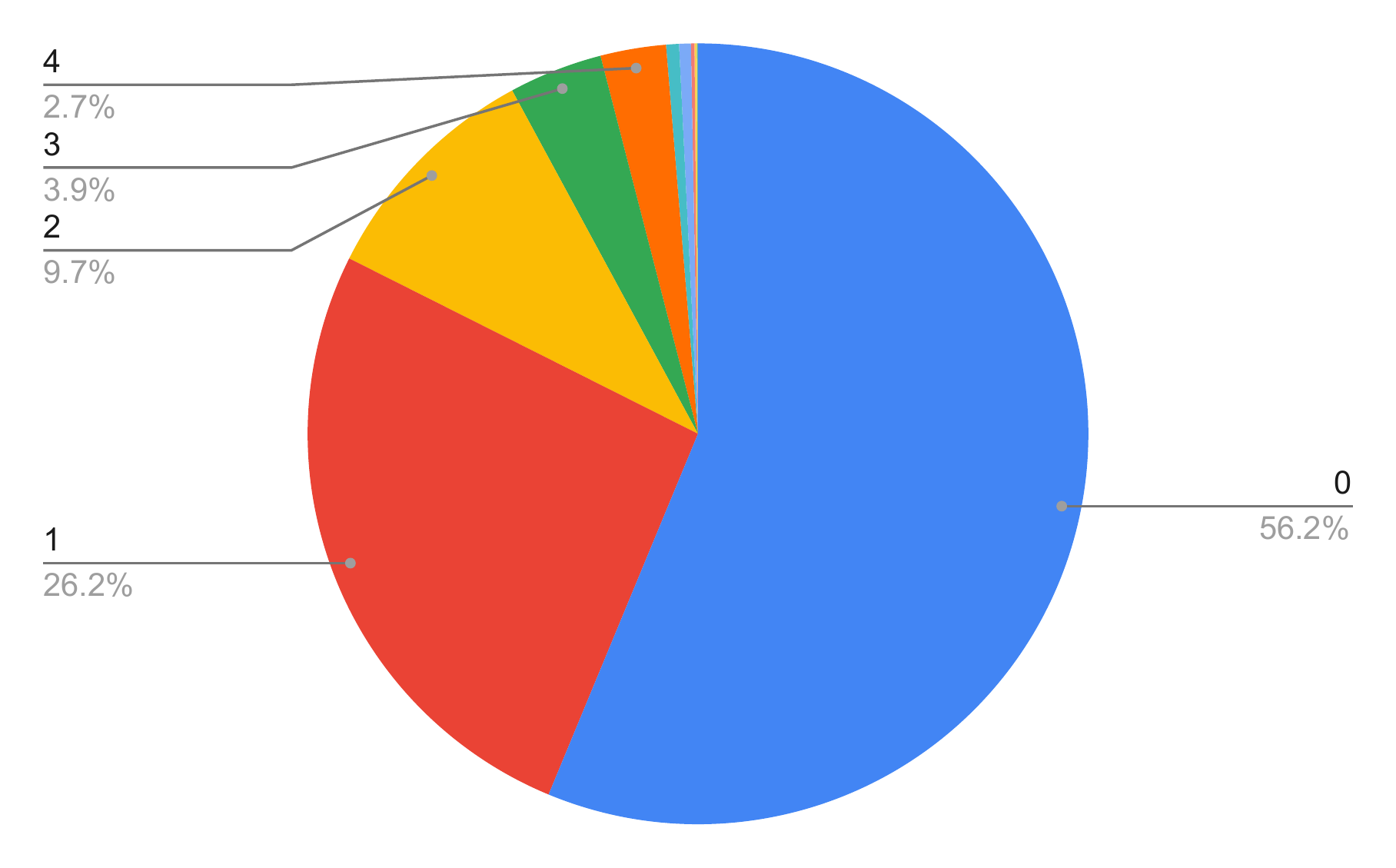}
\caption{Number of cases of ellipsis in 10 questions of a dialog}
\label{fig:ellipsis}
\end{figure}
}

\end{document}